\documentclass[10pt,journal,compsoc]{IEEEtran}

\usepackage{amsmath}
\usepackage{amssymb}
\usepackage{dsfont}
\usepackage{subcaption}
\usepackage{overpic}
\usepackage{mathtools}
\usepackage{pdfpages}
\usepackage{hyperref}
\usepackage[capitalize]{cleveref}

\usepackage[switch]{lineno}

\usepackage{tikz}
\usetikzlibrary{positioning}
\usetikzlibrary{fit}
\usetikzlibrary{backgrounds}
\usetikzlibrary{calc}

\renewcommand{\vec}[1]{\boldsymbol{#1}}
\newcommand{\mat}[1]{\boldsymbol{#1}}
\newcommand{\seq}[1]{{\boldsymbol{#1}}}
\newcommand{\glob}[1]{\boldsymbol{#1}}

\newcommand{\p}{p}
\newcommand{\given}{\mid}
\newcommand{\without}[1]{-#1}

\newcommand{\integralR}[3]{\int_{#3} #1 \,\mathrm{d} #2}

\newcommand{\particle}[1]{#1^{\lbrace q \rbrace}}

\newcommand{\indicatorS}{\mathds{1}}
\newcommand{\ones}[1]{\mathbf{1}^{#1}}
\newcommand{\idMat}{\mathbf{I}}

\newcommand{\reals}{\mathbb{R}}

\newcommand{\eqpunkt}{.}
\newcommand{\eqkomma}{,}

\newcommand{\defeq}{\coloneqq}

\newcommand{\ie}{i.e.}
\newcommand{\eg}{e.g.}
\newcommand{\cf}{c.f.}
\newcommand{\etc}{etc.}

\newcommand{\Categorical}{\textsc{Cat}}
\newcommand{\Dirichlet}{\textsc{Dir}}
\newcommand{\DirMult}{\textsc{DirMult}}
\newcommand{\Gaussian}{\mathcal{N}}
\newcommand{\Exponential}{\textsc{Exp}}

\tikzstyle{var}=[draw, circle, fill=white]
\tikzstyle{obs}=[draw, circle, fill=black!25!white]
\tikzstyle{hyper}=[draw, circle, fill=black, minimum size=0.2cm]
\tikzstyle{fac}=[draw, rectangle, fill=black, minimum size=0.25cm]
\tikzstyle{facDummy}=[rectangle, minimum size=0.25cm]
\tikzstyle{background}=[rectangle, rounded corners=5mm, fill=black!10]

\def \fitspace {6mm}

\def \lw {0.7cm}
\def \ax {3.6cm}
\def \dist {1.4cm}
\def \distl {2.6cm}
\def \off {0.35cm}

\ifCLASSOPTIONcompsoc
  \usepackage[nocompress]{cite}
\else
  \usepackage{cite}
\fi
\ifCLASSINFOpdf
\else
\fi
\begin{document}
\title{A Bayesian Approach to Policy Recognition \\ and State Representation Learning}

\author{Adrian \v{S}o\v{s}i\'{c}, Abdelhak M. Zoubir and Heinz Koeppl%
\IEEEcompsocitemizethanks{\IEEEcompsocthanksitem Adrian \v{S}o\v{s}i\'{c} is a member of the Signal Processing Group and an associate member of the Bioinspired Communication Systems Lab, Technische Universit\"at Darmstadt, Germany. E-mail: adrian.sosic@spg.tu-darmstadt.de %
\IEEEcompsocthanksitem Abdelhak M. Zoubir is the head of the Signal Processing Group, Technische Universit\"at Darmstadt, Germany. E-mail: zoubir@spg.tu-darmstadt.de %
\IEEEcompsocthanksitem Heinz Koeppl is the head of the Bioinspired Communication Systems Lab and a member of the Centre for Cognitive Science, Technische Universit\"at Darmstadt, Germany. E-mail: heinz.koeppl@bcs.tu-darmstadt.de}}%
\IEEEtitleabstractindextext{%
\begin{abstract}
Learning from demonstration (LfD) %
is the %
process of building %
behavioral models of a task
from demonstrations %
provided by an expert.
These models can be used \eg\ for %
system control %
by generalizing the expert demonstrations to previously unencountered situations. Most LfD methods, however, make strong %
assumptions about the expert behavior, \eg\ they assume the existence of a deterministic optimal ground truth policy or %
require %
direct %
monitoring of the expert's controls, which limits their practical use as part of a 
general system identification framework.  %
In this work, we consider the LfD problem %
in a more general setting where we allow for arbitrary stochastic expert %
policies, without %
reasoning about the %
optimality of the demonstrations. %
Following a Bayesian methodology,
we model the full posterior distribution of possible 
expert %
controllers that explain the provided demonstration data. %
Moreover,
we show that %
our methodology can be applied in a nonparametric context to %
infer the complexity of the %
state representation used by 
the expert, and to learn task-appropriate partitionings of the system state space.%
\end{abstract}

\begin{IEEEkeywords}
learning from demonstration, policy recognition, imitation learning, Bayesian %
nonparametric modeling, Markov chain Monte Carlo, Gibbs sampling, distance dependent Chinese restaurant process
\end{IEEEkeywords}}

\maketitle

\IEEEdisplaynontitleabstractindextext
\IEEEpeerreviewmaketitle

\section{Introduction}
\IEEEPARstart{L}{earning from demonstration} (LfD) 
has 
become a viable alternative to classical reinforcement learning  
as a new 
data-driven learning paradigm for building behavioral models based on 
demonstration data. 
By exploiting the domain knowledge provided by an expert demonstrator, LfD-built models can focus on the relevant parts of a system's state space \cite{argall2009survey} and hence avoid the need of tedious exploration steps performed by reinforcement learners, which often require an impractically high number of interactions with the system environment \cite{deisenroth2015gaussian} and always come with the risk of letting the system run into undesired or unsafe states \cite{abbeel2005exploration}. 
In addition to that,
LfD-built models have been shown to outperform the expert in several experiments 
\cite{michie1990cognitive,sammut1992learning,abbeel2010autonomous}.


However, most existing LfD methods come with strong requirements that limit their practical use in real-world scenarios. 
In particular, they often require direct 
monitoring of the expert's controls (\eg\ \cite{sammut1992learning,pomerleau1991efficient,atkeson1997robot}) which is possible only under laboratory-like conditions, or they need to interact with the target system via a simulator, if not by controlling the system directly (\eg\ \cite{abbeel2004apprenticeship}). Moreover, many methods are restricted to problems with 
finite state 
spaces (\eg\ \cite{panella2015nonparametric}), or they compute only 
point estimates of the relevant system parameters without providing any information about their level of confidence (\eg\ \cite{abbeel2004apprenticeship},\cite{sosic2016},\cite{ziebart2008maximum}). Last but not least, 
the expert is typically assumed to follow an optimal deterministic policy (\eg\ \cite{ng2000algorithms}) or to at least approximate one, 
based on some presupposed degree of confidence in the optimal behavior (\eg\ \cite{ramachandran2007bayesian}). While such an assumption may be reasonable in some situations (\eg\ for problems in robotics involving a human demonstrator \cite{argall2009survey}), it is not appropriate in many others, such as in multi-agent environments, where an optimal deterministic policy often does not 
exist \cite{parsons2012game}.
%
%
%
%
In fact, there are 
many situations in which the assumption of a deterministic expert 
behavior is violated. 
In a more general system identification setting, our goal could be, for instance, 
to detect the deviation of an agent's policy from its known nominal behavior, \eg\ for the purpose of fault or fraud detection (note that the term ``expert'' is slightly misleading in this context). 
%
Also, there are situations in which we 
might not 
want to reason about the optimality of the 
demonstrations; for instance, 
when studying the exploration strategy of an agent who tries 
to model  its environment 
(or the reactions of other agents \cite{hindriks2008opponent}) by randomly triggering different events. In all these cases, existing LfD methods 
can at best approximate the behavior of the expert
as they 
presuppose the existence of some underlying deterministic 
ground truth 
policy.



In this work, we present a novel approach to 
LfD 
in order to address the above-mentioned shortcomings of existing 
methods. Central to our work is the problem of \textit{policy recognition}, that is, extracting 
the (possibly 
stochastic and non-optimal) policy of 
a system from observations of its behavior. 
Taking a 
general system identification view on the problem, 
our goal is herein 
to make as few assumptions about the expert 
behavior 
as possible. In particular, 
we consider the whole class of stochastic expert policies,
without ever reasoning about the optimality of the 
demonstrations. As a result of this, our hypothesis space is \textit{not} restricted to a certain class of ground truth policies, such as deterministic 
or softmax policies (\cf\ \cite{ramachandran2007bayesian}). 
This is in contrast to 
inverse reinforcement learning approaches (see \cref{sec:relatedWork}), which interpret the 
observed demonstrations 
as the result of some preceding planing procedure conducted by the expert
which they try to invert. In the above-mentioned case of 
fault detection, for example, such an inversion attempt 
will generally fail since the demonstrated behavior 
can be arbitrarily far from optimal, which renders an explanation of the data in terms of a simple reward function impossible. 

Another advantage of our problem formulation is that 
the resulting inference machinery 
is entirely passive, 
in the sense that 
we require no active control of the target system 
nor 
access to the action sequence performed by the expert.  
Accordingly, our method is applicable to 
a 
broader range of problems than targeted by 
most existing LfD frameworks and can be used for 
system identification in cases where we cannot interact with the target system.
However, 
%
our objective in this paper is 
twofold: we not only attempt to 
answer the question \textit{how} the expert 
performs a given task but also to infer \textit{which information} is used by the expert
to solve it. This knowledge is captured in the form of a joint posterior distribution over possible 
expert state representations and corresponding state controllers.
%
%
As the complexity of the expert's state representation 
is unknown \textit{a~priori}, we 
finally present 
a Bayesian nonparametric approach to 
explore the 
underlying structure of the system space  
based on the available demonstration data.

\subsection{Problem statement}
\label{sec:problemStatement}
Given a set of expert demonstrations 
in the form of a system trajectory $\seq{s}=(s_1, s_2, \ldots, s_T)\in\mathcal{S}^T$ of length $T$, 
where $\mathcal{S}$ denotes the system state space, 
our goal is to determine the latent control policy used by the expert to generate the state sequence.\footnote{ 
The generalization to multiple trajectories 
is straightforward as they 
are conditionally independent given the system parameters.} 
We formalize this problem as a discrete-time decision-making process (\ie\ we assume that the expert executes exactly one control action per 
trajectory state) and adopt the Markov decision process (MDP) formalism \cite{sutton1998reinforcement} as the underlying framework describing the dynamics of our system. 
More specifically, we consider a reduced MDP $(\mathcal{S}, \mathcal{A}, \mathcal{T}, \pi)$ which consists of a 
countable or uncountable system state space $\mathcal{S}$, a 
finite set of actions $\mathcal{A}$ containing $|\mathcal{A}|$ elements, a transition model $\mathcal{T}:\mathcal{S}\times\mathcal{S}\times\mathcal{A}\rightarrow\mathbb{R}_{\geq0}$ 
where $\mathcal{T}(s'\given s,a)$ denotes the probability (density) assigned to the event of reaching state~$s'$ after taking action $a$ in state $s$, and a policy $\pi$ 
modeling the expert's choice of actions.\footnote{
This reduced model is sometimes referred to as an MDP\textbackslash R (see \eg\ \cite{abbeel2004apprenticeship,syed2007game,michini2012bayesian}) to emphasize the nonexistence of a reward function.}  
%
In the following, we assume that the expert policy is 
parametrized by a 
parameter 
$\glob{\omega}\in\Omega$, which we call 
the \textit{global control parameter} of the system, 
and we write 
$\pi(a\given s,\glob{\omega})$,  
$\pi:\mathcal{A}\times\mathcal{S}\times\Omega\rightarrow[0,1]$, to denote the expert's \textit{local policy} (\ie\ the distribution of actions $a$ played by the expert) at any given state $s$ under 
$\glob{\omega}$. 
The set $\Omega$ is called 
the parameter space of the policy, which 
specifies the class of feasible action distributions. The specific form of $\Omega$ will be discussed later. 

%
%
%
Using a parametric description for $\pi$ 
is convenient as it shifts 
the 
recognition task from determining the possibly infinite set of local policies at all states 
in $\mathcal{S}$ 
to inferring the posterior distribution 
$\p(\glob{\omega} \given \seq{s})$,
which contains all information that is relevant for predicting the expert behavior, 
\begin{linenomath*}
\begin{equation*}
\p(a\given s^*,\seq{s}) = \integralR{\pi(a\given s^*,\glob{\omega})\p(\glob{\omega}\given\seq{s})}{\glob{\omega}}{\Omega} \eqpunkt
\end{equation*}
\end{linenomath*}
Herein, $s^*\in\mathcal{S}$ is some arbitrary query point and $\p(a\given s^*,\seq{s})$ is the corresponding predictive action distribution.
Since the local policies are 
coupled through the global control parameter $\glob{\omega}$ as indicated by the above integral equation, inferring 
$\glob{\omega}$ means not only to determine the individual local policies but also their spatial dependencies. Consequently, learning the structure of 
$\glob{\omega}$ from 
demonstration data 
can be also 
interpreted as 
learning a 
suitable 
state representation for the task performed by the expert. This relationship will 
be discussed in detail in the forthcoming sections. 
In \cref{sec:nonparametric}, we further extend this reasoning to a nonparametric policy model whose hypothesis class finally covers all stochastic policies on $\mathcal{S}$. 

%

For the remainder of this paper, we make the common assumptions that the transition model $\mathcal{T}$ as well as the system state space $\mathcal{S}$ and the action set $\mathcal{A}$ are known. The assumption of knowing $\mathcal{S}$ follows naturally because we already assumed that 
we can observe the expert acting in $\mathcal{S}$. In the proposed Bayesian framework, the latter
assumption can be easily relaxed by 
considering 
noisy or 
incomplete trajectory data. However, as this would 
not provide 
additional insights into the main principles of our method, we do not consider such an extension in this work. 

The assumption of knowing the transition 
dynamics $\mathcal{T}$ 
is a simplifying one but 
prevents us from running into model identifiability problems:
%
%
if we do not constrain 
our system transition model 
in some 
reasonable way, any observed state transition in $\mathcal{S}$ could be 
trivially explained 
by a corresponding local adaptation of the assumed transition 
model $\mathcal{T}$ 
and, thus, there would be little 
hope to extract the true expert policy from the demonstration data. 
Assuming a 
fixed transition model is the easiest way to resolve this model ambiguity. However, there are 
alternatives which we leave for future work, 
for example, using a parametrized family of transition models for joint 
inference. 
This extension can be integrated seamlessly into our Bayesian framework and is useful in cases where we can constrain the system dynamics in a natural way, \eg\ when modeling physical processes. 
Also, it should be mentioned that 
we can tolerate deviations from the true system dynamics as long as 
our model $\mathcal{T}$ is 
sufficiently accurate to 
extract 
information about the 
expert action sequence \textit{locally}, 
because our inference algorithm naturally processes the demonstration data piece-wise in the form of one-step state transitions $\{(s_t,s_{t+1})\}$ (see algorithmic details in \cref{sec:parametricPolicyRecognition} and results in \cref{sec:discreteSetting}).
%
%
%
This is in contrast to 
planning-based approaches, where small modeling errors in the dynamics 
can accumulate and yield consistently wrong policy 
estimates \cite{atkeson1997comparison,atkeson1997robot}.

The requirement of knowing the action set $\mathcal{A}$ is 
less stringent: 
if 
$\mathcal{A}$ is unknown \textit{a priori}, we can still 
assume a potentially rich class of actions, 
as long as 
the transition model 
can 
provide the corresponding dynamics (see example in \cref{sec:discreteSetting}). 
For instance, we might be able to provide a model 
which describes 
the movement of a robotic arm even if 
 the maximum torque that can be generated by the system is unknown. 
Figuring out which of the hypothetical actions are actually 
performed by the expert and, more importantly, how they are used in a given context, 
shall be
the task of our inference algorithm.

\subsection{Related work}
\label{sec:relatedWork}
The idea of learning from demonstration 
has now been around for several decades. 
%
Most of the work on LfD has been presented by the robotics community (see \cite{argall2009survey} for a survey), 
but recent advances in the field have triggered 
developments in other research areas, 
such as cognitive science \cite{rothkopf2011preference} and human-machine interaction \cite{pietquin2013inverse}.
%
Depending on the 
setup, 
the problem is referred to as imitation learning \cite{schaal1999imitation}, apprenticeship learning \cite{abbeel2004apprenticeship}, inverse reinforcement learning \cite{ng2000algorithms}, inverse optimal control \cite{dvijotham2010inverse}, preference elicitation \cite{rothkopf2011preference}, plan recognition \cite{charniak1993bayesian} or behavioral cloning \cite{sammut1992learning}. Most LfD models can be categorized 
as intentional models (with inverse reinforcement learning models as the primary example), 
or sub-intentional models (\eg\ behavioral cloning models). 
%
%
%
%
While the latter class 
only 
predicts an agent's behavior via a learned 
policy representation, intentional models (additionally) attempt to capture the agent's beliefs and intentions, \eg\ in the form of a reward function. 
For this reason, intentional models are 
often 
reputed to have better generalization abilities\footnote{The rationale behind this 
is that an agent's intention is always specific to the task being performed and can hence serve as a compact description of it 
\cite{ng2000algorithms}. However, if the 
intention of the agent is misunderstood, 
then also the subsequent generalization step \mbox{will trivially 
fail.}}; however, they typically require a certain amount of task-specific prior knowledge in order to 
resolve the ambiguous relationship between intention and behavior, since there are often many ways to 
solve a certain task \cite{ng2000algorithms}. 
%
%
%
Also, 
albeit being interesting 
from a psychological point of view \cite{rothkopf2011preference}, intentional models 
target a much harder problem than what is actually required in many LfD scenarios. For instance, it is not 
necessary to 
understand an agent's intention if we only wish to 
analyze its behavior locally. 

Answering the question whether or not an 
intention-based modeling of the LfD problem is advantageous, 
is out of the scope of this paper; however, we point to the 
comprehensive discussion 
in \cite{piot2013learning}. Rather, we present 
a hybrid solution containing both intentional and sub-intentional elements. More specifically, our method does not explicitly capture the expert's goals in the form of a reward function but infers a policy model directly from the demonstration data; nonetheless, the presented algorithm learns a task-specific representation of the system state space which encodes the structure of the underlying control problem to facilitate the policy prediction task. 
An early version of this idea can be found 
in \cite{waltz1965heuristic}, 
where the authors proposed
a simple 
method to partition 
a system's state space into a set of so-called \textit{control situations} to learn a global system controller based on a 
small set of informative states. 
However, their 
framework does not incorporate any demonstration data and the proposed partitioning is based on 
heuristics. 
A more sophisticated partitioning approach 
utilizing 
expert demonstrations is shown in \cite{sosic2016}; yet, the proposed expectation-maximization framework applies to deterministic policies and 
finite state 
spaces only. 

The closest methods to ours can be probably found in \cite{michini2012bayesian} and \cite{panella2015nonparametric}.  The authors of \cite{michini2012bayesian} presented a nonparametric 
inverse reinforcement learning approach to cluster the expert data based on a set of learned subgoals encoded in the form of local rewards. 
Unfortunately, 
the 
required subgoal assignments are learned only for the demonstration set and, thus, the algorithm 
cannot be used for action prediction at unvisited states unless it is extended with a non-trivial post-processing step which solves the subgoal assignment problem.
Moreover, the algorithm requires an MDP solver, which 
causes difficulties for systems with uncountable state spaces. The sub-intentional model in \cite{panella2015nonparametric}, on the other hand, can be used to learn a class of finite state controllers directly from the expert demonstrations. 
Like our framework, the algorithm can handle various kinds of uncertainty about the data but, again, 
the proposed approach is limited to discrete settings.
%
%
In the context of reinforcement learning, we further 
point to the work presented in \cite{doshi2015bayesian} 
whose authors follow a nonparametric 
strategy similar to ours, to learn a distribution over predictive state representations for decision-making.

\subsection{Paper outline}
The outline of the paper is as follows: 
%
%
In \cref{sec:parametricPolicyRecognition}, we introduce our parametric policy recognition framework and derive inference algorithms for 
both countable and uncountable state spaces. In \cref{sec:nonparametric}, we consider the policy recognition problem from a nonparametric viewpoint and 
provide insights into the state representation learning problem. 
Simulation results are presented in \cref{sec:results} and we give a conclusion of 
our work in \cref{sec:conclusion}.
In the supplement, we provide additional simulation results, a note on the computational complexity of our model, as well as an in-depth discussion on the issue of marginal invariance and the problem of policy prediction in large states spaces.

\section{Parametric Policy Recognition}
\label{sec:parametricPolicyRecognition}

\subsection{Finite state spaces: the static model}
First, 
let us assume that 
the expert system can be modeled 
on a finite state space $\mathcal{S}$ and let $|\mathcal{S}|$ denote 
its cardinality. 
For notational convenience, we 
represent both 
states and actions by integer values. 
%
%
Starting with the most general case, we assume that the expert executes
an individual control strategy at each possible system state. 
Accordingly, we introduce a set of \textit{local control parameters} or \textit{local controllers} $\{\vec{\theta}_i\}_{i=1}^{|\mathcal{S}|}$ 
by which we describe the expert's choice of actions. 
More specifically, 
we model the executed actions as categorical random variables and 
let the $j\text{th}$ element of $\vec{\theta}_i$ represent the probability 
that the expert chooses action $j$ at state~$i$. Consequently, $\vec{\theta}_i$ lies in the $(|\mathcal{A}|-1)$-simplex, which we 
denote by the symbol $\Delta$ for brevity of notation, \ie\ $\vec{\theta}_i\in\Delta\subseteq\mathbb{R}^{|\mathcal{A}|}$. 
Summarizing 
all 
local control parameters in a single 
matrix, $\glob{\Theta}\in\Omega\subseteq\Delta^{|\mathcal{S}|}$, 
we obtain the \textit{global control parameter} of the system as already introduced 
in \cref{sec:problemStatement}, which compactly 
captures the 
expert behavior. 
Note that we denote the global control parameter here by $\glob{\Theta}$ instead of~$\glob{\omega}$, for reasons that will become clear later.
Each action $a$ 
is thus 
characterized by the 
local 
policy that is induced by the 
control parameter 
of the underlying state, 
\begin{linenomath*}
\begin{equation*}
	\pi(a \given s=i, \glob{\Theta}) = \Categorical(a \given \vec{\theta}_{i}) \eqpunkt 
\end{equation*} 
\end{linenomath*}
For simplicity, we will 
write $\pi(a \given \vec{\theta}_i)$ 
since the state 
information is used only to select the appropriate 
local controller. 

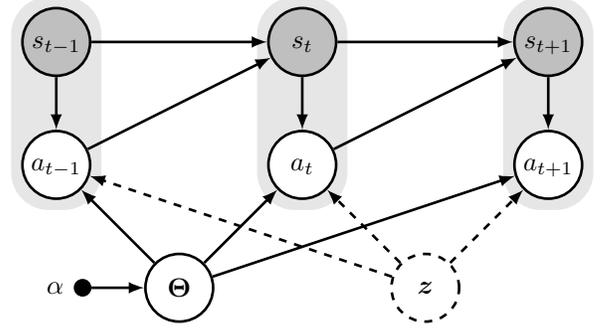
\begin{figure}
	\centering
	\begin{tikzpicture}[inner sep=0cm, minimum size = 0.9cm, line width = 1pt, text depth = 0ex, -latex]
		\node (state1) [obs] {$s_{t-1}$};
		\node (stateDummy1) [right = 0.7cm of state1] {};
		\node (state2) [obs, right=0.7cm of stateDummy1] {$s_{t}$};
		\node (stateDummy2) [right = 0.7cm of state2] {};
		\node (state3) [obs, right=0.7cm of stateDummy2] {$s_{t+1}$};
		\node (action1) [var, below=0.7cm of state1] {$a_{t-1}$};
		\node (action2) [var, below=0.7cm of state2] {$a_{t}$};
		\node (action3) [var, below=0.7cm of state3] {$a_{t+1}$};
		\node (actionDummy1) [below = 0.7cm of stateDummy1] {};
		\node (actionDummy2) [below = 0.7cm of stateDummy2] {};
		\node (policy) [var, below=0.7cm of actionDummy1] {$\glob{\Theta}$};
		\node (alpha) [hyper, left=0.7cm of policy, label={[xshift=0.22cm]left:$\alpha$}] {};
		\node (indicator) [var, dashed, below=0.7cm of actionDummy2] {$\seq{z}$};
		
		\draw (state1) to (state2);
		\draw (state2) to (state3);
		\draw (state1) to (action1);
		\draw (state2) to (action2);
		\draw (state3) to (action3);
		\draw (policy) to (action1);
		\draw (policy) to (action2);
		\draw (policy) to (action3);
		\draw (alpha) to (policy);
		\draw [dashed] (indicator) to (action1);
		\draw [dashed] (indicator) to (action2);
		\draw [dashed] (indicator) to (action3);
		\draw (action1) to (state2);
		\draw (action2) to (state3);	

		\begin{pgfonlayer}{background}
			\node [background, fit={($(state1)+(\fitspace,\fitspace)$) ($(action1)+(-\fitspace,-\fitspace)$)}] {};
			\node [background, fit={($(state2)+(\fitspace,\fitspace)$) ($(action2)+(-\fitspace,-\fitspace)$)}] {};
			\node [background, fit={($(state3)+(\fitspace,\fitspace)$) ($(action3)+(-\fitspace,-\fitspace)$)}] {};
		\end{pgfonlayer}		
	\end{tikzpicture}
	\caption{Graphical model 
of the policy recognition framework. 
The 
underlying
dynamical structure is that of an MDP 
whose global control parameter $\glob{\Theta}$ is 
treated as a random variable with prior distribution parametrized by $\alpha$. The indicator 
node $\seq{z}$ is used for the clustering model in \cref{sec:clusteringApproach}. Observed variables are 
highlighted 
in gray.}
	\label{fig:graphicalModel}
\end{figure}

Considering a finite set of actions, 
it is convenient to place a symmetric Dirichlet 
prior on the local control parameters, 
\begin{linenomath*}
\begin{equation*}
	\p_{\theta}(\vec{\theta}_i \given \alpha) = \Dirichlet(\vec{\theta}_i \given \alpha \cdot \ones{|\mathcal{A}|}) \eqkomma 
\end{equation*}
\end{linenomath*}
which forms the conjugate distribution to the categorical distribution over actions.
%
%
Here, $\ones{|\mathcal{A}|}$ denotes the 
vector of all ones of length $|\mathcal{A}|$. 
The prior is itself parametrized by a concentration parameter $\alpha$ which 
can be further described by a hyperprior $\p_\alpha(\alpha)$, giving rise to a Bayesian hierarchical model. 
For simplicity, we assume that the value of $\alpha$ is fixed for the remainder of this paper, but 
the extension to a full Bayesian treatment is straightforward.
The 
joint distribution of all remaining model variables 
is, therefore, given as
\begin{linenomath*}
\begin{align}
	\p(\seq{s},\seq{a},\glob{\Theta} &\given \alpha) = \p_1(s_1)\prod_{i=1}^{|\mathcal{S}|} \p_\theta(\vec{\theta}_i \given \alpha) \ \ldots \label{eq:staticJoint}\\
	&\ldots\ \times \prod_{t=1}^{T-1} \mathcal{T}(s_{t+1} \given s_t, a_t) \pi(a_t \given \vec{\theta}_{s_t}) \eqkomma \nonumber
\end{align}
\end{linenomath*}
where $\seq{a}=(a_1, a_2, \ldots, a_{T-1})$ denotes the latent action sequence taken by the expert and $\p_1(s_1)$ is the initial state distribution of the system. 
Throughout the rest of the paper, we refer to this model as the \textit{static model}.
The corresponding graphical visualization 
is depicted in \cref{fig:graphicalModel}. 
%
%



\subsubsection{Gibbs sampling}
\label{sec:GibbsSamplerStatic}
Following a Bayesian methodology, our goal is to determine the posterior distribution $\p(\glob{\Theta} \given \seq{s}, \alpha)$, which contains all information necessary to make predictions about the expert behavior. 
For the static model in \cref{eq:staticJoint}, the 
required marginalization of the 
latent action sequence~$\seq{a}$ can be computed efficiently because the 
joint distribution factorizes over time instants. For 
the extended models presented in later sections, however, 
a direct marginalization becomes computationally 
intractable due to the exponential growth of latent variable configurations. As a solution to this problem, we 
follow a sampling-based inference strategy which is later on generalized to more complex settings.

For the simple model described above, we 
%
%
first 
approximate the joint posterior distribution $\p(\glob{\Theta},\seq{a} \given \seq{s}, \alpha)$ over both controllers and actions using a finite number of $Q$ samples, and then 
marginalize over $\seq{a}$ in a second step, 
\begin{linenomath*}
\begin{align}
	\p(\glob{\Theta} &\given \seq{s}, \alpha) = \sum_\seq{a} \p(\glob{\Theta}, \seq{a} \given \seq{s}, \alpha) \label{eq:MCapproxGibbsStatic}\\
	&\approx \sum_\seq{a} \left(\frac{1}{Q} \sum_{q=1}^Q \delta_{\particle{\glob{\Theta}}\particle{\seq{a}}}(\glob{\Theta},\seq{a})\right) = \frac{1}{Q} \sum_{q=1}^Q \delta_{\particle{\glob{\Theta}}}(\glob{\Theta}) \eqkomma \nonumber
\end{align}
\end{linenomath*}
where $(\particle{\glob{\Theta}},\particle{\seq{a}}) \sim \p(\glob{\Theta}, \seq{a} \given \seq{s}, \alpha)$, and $\delta_x(\cdot)$ denotes 
Dirac's delta function centered at $x$. 
This two-step approach
gives rise to a simple inference procedure since
the joint samples $\lbrace(\particle{\glob{\Theta}},\particle{\seq{a}})\rbrace_{q=1}^Q$  can be easily obtained from a Gibbs sampling scheme,
\ie\ by sampling
iteratively from the following two 
conditional distributions, 
\begin{linenomath*}
\begin{align*}
\p(a_t \given \seq{a}_{\without{t}}, \seq{s}, \glob{\Theta}, \alpha) &\propto \mathcal{T}(s_{t+1} \given s_t, a_t) \pi(a_t \given \vec{\theta}_{s_t}) \eqkomma \\
\p(\vec{\theta}_i \given \vec{\Theta}_{\without{i}}, \seq{s}, \seq{a}, \alpha) &\propto \p_\theta(\vec{\theta}_i \given \alpha) \prod_{t:s_t=i} \pi(a_t \given \vec{\theta}_i) \eqpunkt
\end{align*}
\end{linenomath*}
Herein, $\seq{a}_{\without{t}}$ and $\glob{\Theta}_{\without{i}}$ refer to all 
actions/controllers except $a_t$ and $\vec{\theta}_i$, respectively. The latter of the two 
expressions reveals that, in order to sample $\vec{\theta}_i$, we need to consider only those actions 
played at the corresponding state $i$. Furthermore, the first expression shows that, given $\glob{\Theta}$, all actions $\{a_t\}$ can be sampled independently of each other. Therefore, inference can be done in parallel for all $\vec{\theta}_i$. This can be also seen from the posterior distribution of 
the global control parameter, which factorizes over states,
\begin{linenomath*}
\begin{equation}
\p(\glob{\Theta} \given \seq{s}, \seq{a}, \alpha) \propto \prod_{i=1}^{|\mathcal{S}|} \p_\theta(\vec{\theta}_i \given \alpha) \prod_{t:s_t=i} \pi(a_t \given \vec{\theta}_i) \eqpunkt
\label{eq:staticControlPosterior}
\end{equation}
\end{linenomath*}
From the conjugacy of $\p_\theta(\vec{\theta}_i \given \alpha)$ and $\pi(a_t \given \vec{\theta}_i)$, it follows that the posterior over $\vec{\theta}_i$ is again Dirichlet distributed with updated concentration parameter. In particular, denoting by $\phi_{i,j}$ the number of times that action $j$ is played at state $i$ 
for the current assignment of actions $\seq{a}$, 
\begin{linenomath*}
\begin{equation}
	\phi_{i,j} \defeq \sum_{t:s_t=i} \indicatorS(a_t=j) \eqkomma
	\label{eq:definitionStateActionCounts}
\end{equation}
\end{linenomath*}
and by collecting these 
quantities in the form of 
vectors, \ie\ $\vec{\phi}_i\defeq[\phi_{i,1},\ldots,\phi_{i,|\mathcal{A}|}]$, 
%
we can rewrite \cref{eq:staticControlPosterior} as
\begin{linenomath*}
\begin{equation}
	\p(\glob{\Theta} \given \seq{s}, \seq{a}, \alpha) = \prod_{i=1}^{|\mathcal{S}|} \Dirichlet(\vec{\theta}_i \given \vec{\phi}_{i} + \alpha \cdot \ones{|\mathcal{A}|}) \eqpunkt
	\label{eq:staticControlPosteriorDir}
\end{equation}
\end{linenomath*}

\subsubsection{Collapsed Gibbs sampling}
\label{sec:GibbsCollapsedFinite}
Choosing a Dirichlet distribution as prior model for the local 
controllers is convenient as it allows us to  arrive at analytic expressions for the conditional distributions that are required 
to run the Gibbs sampler. As an alternative, 
we can 
exploit the conjugacy property of $\p_\theta(\vec{\theta}_i \given \alpha)$ and $\pi(a_t \given \vec{\theta}_i)$ to marginalize out the 
control parameters during the sampling process, 
giving rise to a collapsed 
sampling scheme. Collapsed sampling 
is advantageous in two different respects: 
first, it reduces the total 
number of variables to be sampled and, 
hence,
the number of computations required per Gibbs iteration; second, it increases the mixing speed of the underlying Markov chain 
that governs the sampling process, reducing the correlation of the obtained samples and, with it, the variance of the resulting 
policy estimate.

Formally, collapsing means that we no longer approximate the joint distribution $\p(\glob{\Theta},\seq{a} \given \seq{s}, \alpha)$ as done in \cref{eq:MCapproxGibbsStatic}, but instead 
sample from the marginal density $\p(\seq{a} \given \seq{s}, \alpha)$,
\begin{linenomath*}
\begin{align}
	\p(\glob{\Theta} \given \seq{s}, \alpha) &= \sum_\seq{a} \p(\glob{\Theta} \given \seq{s}, \seq{a}, \alpha) \p(\seq{a} \given \seq{s}, \alpha) \nonumber \\
	&\approx \sum_\seq{a} \p(\glob{\Theta} \given \seq{s}, \seq{a}, \alpha) \left( \frac{1}{Q} \sum_{q=1}^Q \delta_{\particle{\seq{a}}}(\seq{a}) \right) \nonumber \\
	&= \frac{1}{Q} \sum_{q=1}^Q \p(\glob{\Theta} \given \seq{s}, \particle{\seq{a}}, \alpha) \eqkomma \label{eq:DirichletMixtureCollapsed}
\end{align}
\end{linenomath*}
where $\particle{\seq{a}} \sim \p(\seq{a} \given \seq{s}, \alpha)$. In contrast to the previous 
approach, 
the target distribution is no longer 
represented by a sum of Dirac 
measures but 
described 
by a 
product of Dirichlet mixtures (compare \cref{eq:staticControlPosteriorDir}). 
%
%
The required samples $\lbrace \particle{\seq{a}} \rbrace$ can be obtained 
from a collapsed Gibbs sampler 
with
\begin{linenomath*}
\begin{align*}
	\p(&a_t \given \seq{a}_{\without{t}}, \seq{s}, \alpha) 
	\propto \integralR{\p(\seq{s},\seq{a},\glob{\Theta} \given \alpha)}{\glob{\Theta}}{\Delta^{|\mathcal{S}|}} \\
	&\propto \mathcal{T}(s_{t+1} \given s_t, a_t) \integralR{\p_\theta(\vec{\theta}_{s_t} \given \alpha) \prod_{t':s_{t'}=s_t} \pi(a_{t'} \given \vec{\theta}_{s_t})}{\vec{\theta}_{s_t}}{\Delta} \eqpunkt
\end{align*}
\end{linenomath*}
It turns out that the above distribution  
provides an easy sampling mechanism since the integral part, when viewed as a function of action $a_t$ only, can be identified as the conditional of a Dirichlet-multinomial distribution. 
This distribution is then reweighted by the likelihood $\mathcal{T}(s_{t+1} \given s_t, a_t)$ of the observed transition. The final (unnormalized) weights of the resulting categorical distribution are hence given as
\begin{linenomath*}
\begin{equation}
	\p(a_t=j \given \seq{a}_{\without{t}}, \seq{s}, \alpha) \propto \mathcal{T}(s_{t+1} \given s_t, a_t=j) \cdot (\varphi_{t,j} + \alpha) \eqkomma
	\label{eq:conditionalActionsDirichletCollapsed}
\end{equation}
\end{linenomath*}
where  
$\varphi_{t,j}$ 
counts the number of occurrences of action $j$ among all actions in $\seq{a}_{\without{t}}$ played at the same state as $a_t$ (that is, $s_t$). 
Explicitly,
\begin{linenomath*}
\begin{equation*}
	\varphi_{t,j} \defeq \sum_{\substack{t':s_{t'}=s_t \\ t' \neq t}} \indicatorS(a_{t'}=j) \eqpunkt
\end{equation*}
\end{linenomath*}
Note that these values can be also 
expressed in terms of the sufficient statistics introduced in the last section,
\begin{linenomath*}
\begin{equation*}
\varphi_{t,j} = \phi_{s_t,j} - \indicatorS(a_t=j) \eqpunkt
\end{equation*}
\end{linenomath*}
As before, actions played at different states may be sampled independently of each other 
because they 
are 
generated by different local controllers. Consequently, inference about $\glob{\Theta}$ again decouples for all states.


\subsection{Towards large state spaces: a clustering approach}
\label{sec:clusteringApproach}
While the methodology introduced so far provides a means to solve the policy recognition problem 
in finite state spaces, the presented approaches 
quickly become infeasible for large 
spaces as, in the continuous limit, the number of parameters to be learned (\ie\ the size of $\glob{\Theta}$) will grow unbounded. 
In that sense, 
the presented methodology is prone to overfitting 
because, for larger problems, we will never have enough demonstration data to sufficiently cover the whole system
state space. 
In particular, the static model makes no assumptions about the structure of $\glob{\Theta}$ but treats all local policies 
separately (see \cref{eq:staticControlPosteriorDir}); hence, we are not able to 
generalize the demonstrated behavior to regions of the state space that 
are not directly visited by the expert.  Yet, we would certainly like to 
predict the expert 
behavior also at states for which there is no trajectory data available. Moreover, we should 
expect 
a 
well-designed model to produce increasingly accurate predictions at regions closer to the observed 
trajectories (with the precise definition of ``closeness'' being left open for the moment). 

\begin{figure}
	\centering
	\includegraphics[width=6cm]{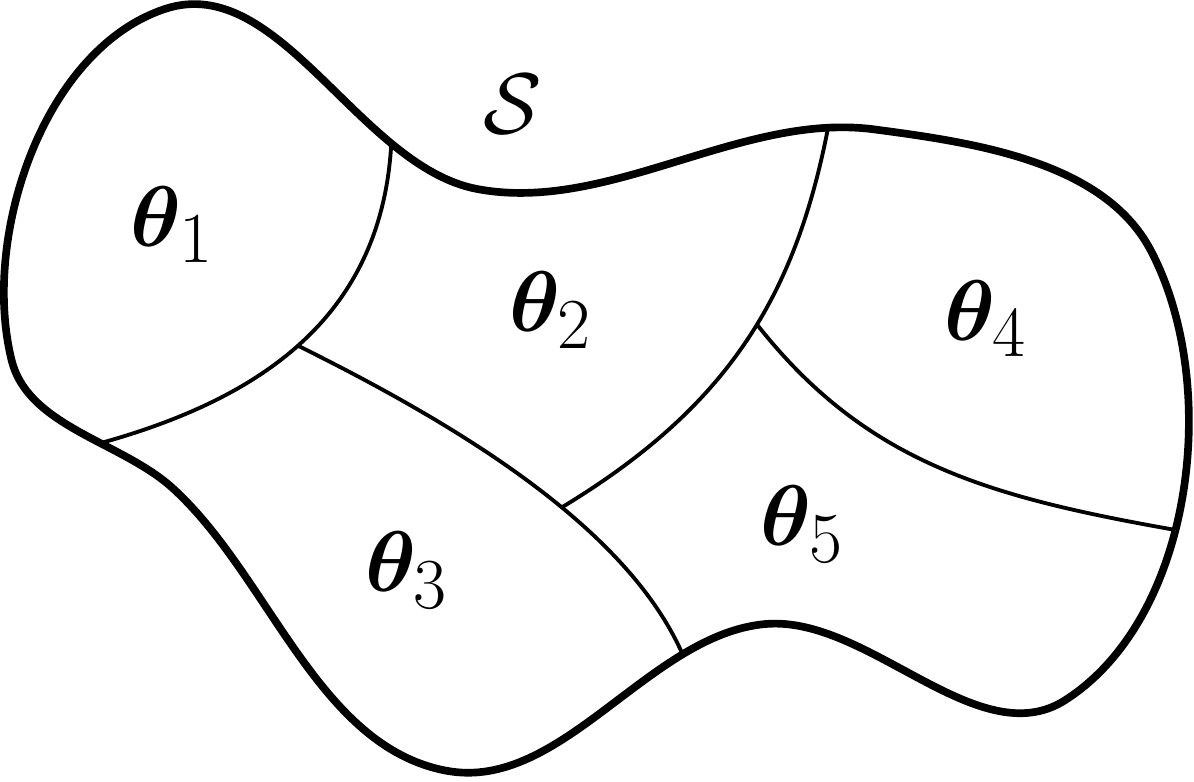}
	\caption{Schematic illustration of the clustering model. The state space $\mathcal{S}$ is partitioned into a 
	set of clusters $\lbrace\mathcal{C}_k\rbrace$, 
	each governed by its 
	own local control parameter $\vec{\theta}_k$. }
	\label{fig:schematicPartitioning}
\end{figure}

A simple way to counteract the overfitting problem, in general, is to restrict the complexity of a model by limiting the number of its free parameters. In our case, we can avoid the parameter space to grow unbounded by 
considering only 
a finite number of local policies 
that need to be shared between the states.
The underlying assumption is that,
at each state, 
the expert selects 
an action 
according to one of $K$ local policies, with corresponding control 
parameters $\lbrace \vec{\theta}_k \rbrace_{k=1}^K$. Accordingly, we introduce a set of indicator or cluster assignment variables, $\lbrace z_i \rbrace_{i=1}^{|\mathcal{S}|}$, $z_i\in\{1,\ldots,K\}$, 
which 
map the states to their local controllers (\cref{fig:graphicalModel}). Obviously, such an assignment implies a partitioning of the state space (\cref{fig:schematicPartitioning}), 
resulting in the following $K$ clusters, 
\begin{linenomath*}
\begin{equation*}
\mathcal{C}_k \defeq \lbrace i : z_{i} = k \rbrace, \quad  k\in\lbrace 1,\ldots,K\rbrace \eqpunkt
\end{equation*}
\end{linenomath*}

Although we motivated the clustering of states by the problem of overfitting,  
partitioning a system's space 
is not only convenient from 
a statistical point of view; 
mapping the 
inference problem down 
to a lower-dimensional space is also reasonable 
for practical reasons as 
we are typically interested in understanding an agent's behavior on a certain 
task-appropriate scale. 
The following paragraphs discuss these reasons in detail:

	\vspace{0.5em}
	\noindent $\bullet$  In practice, the observed trajectory data will always be noisy since we can take our measurements only up to 
	a certain finite precision. 
	Even though we do not explicitly consider 
	observation noise in this paper, 
	clustering 
	the data appears reasonable in order to robustify the model against 
	small 
	perturbations in our observations.

	\vspace{0.5em}
	\noindent $\bullet$ 
	Considering the 
	LfD problem from a control perspective, the complexity of 
	subsequent planning steps can be 
	potentially reduced if 
	the 
	system 
	dynamics can be approximately described on a lower-dimensional manifold of the state space, meaning that the system behavior can be well represented by a smaller set of informative states (\cf\ finite state controllers \cite{meuleau1999learning}, control situations \cite{waltz1965heuristic}). 
	The LfD problem can then be 
	interpreted as the problem of learning a (near-optimal) 
	controller based on a small set of local policies that together provide a good approximation of the global agent behavior. What remains is the question how we can find such a representation. 
	The clustering approach described above offers one possible solution to this problem.
	
	\vspace{0.5em}
	\noindent $\bullet$ Finally, 
	in any real setup, it is 
	reasonable to assume that 
	the expert itself can only execute a finite-precision policy due to its own limited sensing abilities of the system state space. Consequently, the demonstrated 
	behavior is going to be optimal only up to a certain finite precision because the agent 
	is generally not able to discriminate between arbitrary small differences of states. An interesting question in this context is whether we can infer the 
	underlying state representation of the expert 
	by observing its reactions to the environment in the form of the resulting state trajectory. We will discuss this issue in detail in \cref{sec:nonparametric}.
	\vspace{0.5em}

\noindent By introducing the cluster assignment variables $\lbrace z_i\rbrace$, the joint distribution in \cref{eq:staticJoint} changes into
\begin{linenomath*}
\begin{align}
	\p(\seq{s},\seq{a},\seq{z},\,&\glob{\Theta} \given \alpha) = \p_1(s_1) \prod_{k=1}^{K} \p_\theta(\vec{\theta}_k \given \alpha)\ \ldots 	\label{eq:fullClusterModel} \\ &\hspace{4mm} \ldots\ \times\prod_{t=1}^{T-1} \mathcal{T}(s_{t+1} \given s_t, a_t) \pi(a_t \given \vec{\theta}_{z_{s_t}}) \p_z(\seq{z}) \eqkomma \nonumber
\end{align}
\end{linenomath*}
where $\seq{z} = (z_1, z_2, \ldots, z_{|\mathcal{S}|})$ denotes the collection of all indicator variables and $p_z(\seq{z})$ is the corresponding prior distribution 
to be further discussed in \cref{sec:priorModels}. Note that the static model can be recovered as a special case of 
the above when each state describes its own cluster, \ie\ by setting $K=|\mathcal{S}|$ and fixing $z_i=i$ (hence the name \textit{static}).


In contrast to the static model, we now require both the indicator $z_i$ and the corresponding 
control parameter $\vec{\theta}_{z_i}$ in order to characterize the expert's behavior at a given state $i$. Accordingly, the global control parameter of the model is $\glob{\omega}=(\glob{\Theta},\seq{z})$ with 
underlying parameter space $\Omega \subseteq 
\Delta^K\times\lbrace 1,\ldots,K \rbrace^{|\mathcal{S}|}$ (see \cref{sec:problemStatement}), and our target distribution becomes $\p(\glob{\Theta}, \seq{z} \given \seq{s}, \alpha)$. In what follows, we derive the Gibbs and the collapsed Gibbs sampler as mechanisms for approximate inference in this setting.

\subsubsection{Gibbs sampling}
\label{sec:GibbsClustering}
As shown by the following equations, the expressions for the conditional distributions over actions and controllers 
take a similar form to those of the static model. Here, the only difference is that we no longer group the actions by their states but according to their generating local policies or, equivalently, the clusters $\lbrace \mathcal{C}_k \rbrace$, 
\begin{linenomath*}
\begin{alignat*}{2}
	&\p(a_t \given \seq{a}_{\without{t}}, \seq{s}, \seq{z}, \glob{\Theta}, \alpha) \ &&\propto \ \mathcal{T}(s_{t+1} \given s_t, a_t) \cdot \pi(a_t \given \vec{\theta}_{z_{s_t}}) \eqkomma \\
	&\p(\vec{\theta}_k \given \glob{\Theta}_{\without{k}}, \seq{s}, \seq{a}, \seq{z}, \alpha) \ &&\propto \ \p_\theta(\vec{\theta}_k \given \alpha) \prod_{t:z_{s_t}=k} \pi(a_t \given \vec{\theta}_k) \\
	 &  &&= \ \p_\theta(\vec{\theta}_k \given \alpha) \prod_{t:s_t\in\mathcal{C}_k} \pi(a_t \given \vec{\theta}_k) \eqpunkt
\end{alignat*}
\end{linenomath*}
The latter expression again takes the form of a Dirichlet distribution with updated concentration parameter,
\begin{linenomath*}
\begin{equation*}
	\p(\vec{\theta}_k \given \glob{\Theta}_{\without{k}}, \seq{s}, \seq{a}, \seq{z}, \alpha) = \Dirichlet(\vec{\theta}_k \given \vec{\xi}_{k}+\alpha \cdot \ones{|\mathcal{A}|}) \eqkomma
\end{equation*}
\end{linenomath*}
where $\vec{\xi}_k\defeq[\xi_{k,1}\ ,\ \ldots\ ,\ \xi_{k,|\mathcal{A}|}]$, and $\xi_{k,j}$ denotes the number of times that action $j$ is played at states belonging to cluster $\mathcal{C}_k$ in the current assignment of $\seq{a}$. Explicitly,
\begin{linenomath*}
\begin{equation}
	\xi_{k,j} \defeq \sum_{t:z_{s_t}=k} \indicatorS(a_t=j) = \sum_{i\in\mathcal{C}_k} \sum_{t:s_t=i} \indicatorS(a_t=j) \eqkomma
	\label{eq:definitionPolicyActionCounts}
\end{equation}
\end{linenomath*}
which is nothing but the sum of the $\phi_{i,j}$'s of the corresponding states,
\begin{linenomath*}
\begin{equation*}
	\xi_{k,j} = \sum_{i\in\mathcal{C}_k} \phi_{i,j} \eqpunkt
\end{equation*}
\end{linenomath*}
In addition to the actions and control parameters, we now also need to sample the indicators $\lbrace z_i \rbrace$, 
whose conditional distributions can be expressed in terms of the corresponding prior 
model and the likelihood of the triggered actions,
\begin{linenomath*}
\begin{equation}
 	\p(z_i \given \seq{z}_{\without{i}}, \seq{s}, \seq{a}, \glob{\Theta}, \alpha) \ \propto \ \p(z_i \given \seq{z}_{\without{i}}) \prod_{t:s_t=i} \pi(a_t \given \vec{\theta}_{z_i}) \eqpunkt
 	\label{eq:condIndicatorFiniteGibbs}
\end{equation}
\end{linenomath*}


\subsubsection{Collapsed Gibbs sampling}
\label{sec:GibbsCollapsedClustering}
As before, we derive the collapsed Gibbs 
sampler 
by marginalizing out the control parameters,
\begin{linenomath*}
\begin{align}
	&\p(z_i \given \seq{z}_{\without{i}}, \seq{s}, \seq{a}, \alpha) \propto \integralR{\p(\seq{s},\seq{a},\seq{z},\glob{\Theta} \given \alpha)}{\glob{\Theta}}{\Delta^K} 	\label{eq:condIndicatorFiniteCollapsed} \\ 
	&\propto \p(z_i \given \seq{z}_{\without{i}}) \, \integralR{\prod_{k=1}^K \p_\theta(\vec{\theta}_k \given \alpha) \prod_{t=1}^{T-1} \pi(a_t \given \vec{\theta}_{z_{s_t}})}{\glob{\Theta}}{\Delta^K} \nonumber \\
	&\propto \p(z_i \given \seq{z}_{\without{i}}) \, \integralR{\prod_{k=1}^K \p_\theta(\vec{\theta}_k \given \alpha) \prod_{i'=1}^{|\mathcal{S}|} \prod_{t:s_t=i'} \pi(a_t \given \vec{\theta}_{z_{i'}})}{\glob{\Theta}}{\Delta^K} \nonumber \\
	&\propto \p(z_i \given \seq{z}_{\without{i}}) \, \integralR{\prod_{k=1}^K \p_\theta(\vec{\theta}_k \given \alpha) \prod_{i':z_{i'}=k} \prod_{t:s_t=i'} \pi(a_t \given \vec{\theta}_k)}{\glob{\Theta}}{\Delta^K} \nonumber \\
	&\propto \p(z_i \given \seq{z}_{\without{i}}) \, \prod_{k=1}^K \left( \integralR{\p_\theta(\vec{\theta}_k \given \alpha) \prod_{t:s_t\in\mathcal{C}_k} \pi(a_t \given \vec{\theta}_k)}{\vec{\theta}_k}{\Delta} \right) \eqpunkt \nonumber
\end{align}
\end{linenomath*}
Here, we first grouped the actions by their associated states and then grouped the states themselves by 
the clusters $\lbrace \mathcal{C}_k \rbrace$. Again,
this distribution 
admits an easy sampling mechanism as it takes the form of a product of Dirichlet-multinomials, reweighted by the conditional prior distribution over indicators. In particular, we observe that all actions played at some state $i$ appear in exactly one of the $K$ integrals of the last equation. In other words, by changing the value of $z_i$ (\ie\ by assigning state $i$ to another cluster), only two of the involved integrals are affected: the one belonging to the previously assigned cluster, and the one 
of the new cluster. Inference about the value of $z_i$ can thus be carried out using the following two sets of sufficient statistics:
\begin{itemize}
	\item $\phi_{i,j}$: the number of actions $j$ played at state $i,$ 
	\item $\psi_{i,j,k}$: the number of actions $j$ played at states assigned to 
	cluster $\mathcal{C}_k$, excluding state $i$.
\end{itemize}
The $\phi_{i,j}$'s are 
the same as in \cref{eq:definitionStateActionCounts} and their definition is repeated here just as a reminder. For the $\psi_{i,j,k}$'s, on the other hand, we find the following explicit expression,
\begin{linenomath*}
\begin{equation*}
	\psi_{i,j,k} \defeq \sum_{\substack{i'\in\mathcal{C}_k \\ i'\neq i}} \sum_{t:s_t=i'} \indicatorS(a_t=j) \eqkomma
\end{equation*}
\end{linenomath*}
which can be also written in terms of the statistics used for the ordinary Gibbs sampler,
\begin{linenomath*}
\begin{equation*}
  \psi_{i,j,k} = \xi_{k,j} - \indicatorS(i\in\mathcal{C}_k) \cdot \phi_{i,j} \eqpunkt
\end{equation*}
\end{linenomath*}
By collecting these quantities in 
a 
vector, \ie\ $\vec{\psi}_{i,k}\defeq[\psi_{i,1,k},\ldots,\psi_{i,|\mathcal{A}|,k}]$, we end up with the following simplified expression,
\begin{linenomath*}
\begin{align*}
	\p(z_i=k &\given \seq{z}_{\without{i}}, \seq{s}, \seq{a}, \alpha) \ \propto \ \p(z_i=k \given \seq{z}_{\without{i}}) \ \ldots \\
	&\ldots \ \times \prod_{k'=1}^K \DirMult(\vec{\psi}_{i,k'} + \indicatorS(k'=k) \cdot \vec{\phi}_i \given \alpha) \eqpunkt
\end{align*}
\end{linenomath*}
Further, we obtain the following result for the conditional distribution of action $a_t$,
\begin{linenomath*}
\begin{align*}
	\p(a_t &\given \seq{a}_{\without{t}}, \seq{s}, \seq{z}, \alpha) \ \propto \ \mathcal{T}(s_{t+1} \given s_t, a_t) \ \ldots \\
	&\ldots \ \times \integralR{\p_\theta(\vec{\theta}_{z_{s_t}} \given \alpha) \prod_{t':z_{s_{t'}}=z_{s_t}} \pi(a_{t'} \given \vec{\theta}_{z_{s_t}})}{\vec{\theta}_{z_{s_t}}}{\Delta} \eqpunkt
\end{align*}
\end{linenomath*}
By introducing the sufficient statistics $\lbrace{\vartheta}_{t,j}\rbrace$, which count the number of occurrences of action $j$ among all states that are currently 
assigned to the same cluster as $s_t$ (\ie\ the cluster 
$\mathcal{C}_{z_{s_t}}$), excluding $a_t$ itself,
\begin{linenomath*}
\begin{equation*}
	{\vartheta}_{t,j} \defeq \sum_{\substack{t':z_{s_{t'}}=z_{s_t} \\ t' \neq t}} \indicatorS({a}_t=j)\eqkomma
\end{equation*}
\end{linenomath*}
we finally arrive at the following expression,
\begin{linenomath*}
\begin{equation*}
	\p(a_t=j \given \seq{a}_{\without{j}}, \seq{s}, \seq{z}, \alpha) \propto ({\vartheta_{t,j}} + \alpha) \cdot \mathcal{T}(s_{t+1} \given s_t, a_t = j) \eqpunkt
\end{equation*}
\end{linenomath*}
As for the static model, we can 
establish a relationship between the statistics used for the ordinary and the collapsed sampler,
\begin{linenomath*}
\begin{equation*}
	{\vartheta}_{t,j} = {\xi}_{z_{s_t},j} - \indicatorS({a}_t=j) \eqpunkt
\end{equation*}
\end{linenomath*}

\subsubsection{Prior models}
\label{sec:priorModels}
In order to complete our model, we 
need to specify a prior distribution over indicator variables $\p_z(\seq{z})$. 
The following paragraphs present three such candidate models: 
\\[\baselineskip]
\textit{Non-informative prior} \\
The simplest of all prior models is the non-informative prior over partitionings, reflecting the assumption that, \textit{a priori}, all cluster assignments are equally likely and that the indicators $\lbrace z_i\rbrace$ are mutually independent. In this case, $\p_z(\seq{z})$ is constant and, hence, the term $\p(z_i \given \seq{z}_{\without{i}})$ in \cref{eq:condIndicatorFiniteGibbs} and \cref{eq:condIndicatorFiniteCollapsed} disappears, 
so that the 
conditional distribution of indicator $z_i$ 
becomes directly proportional to the likelihood of the inferred action sequence.  \\[\baselineskip]
\textit{Mixing prior} \\
Another simple yet expressive prior can be 
realized by the (finite) Dirichlet mixture model. Instead of assuming that the indicator variables are 
independent, 
the model uses a 
set of mixing coefficients $\vec{q}=[q_1, \ldots, q_K]$, 
where $q_k$ 
represents the prior probability that an indicator variable takes on value~$k$. The mixing coefficients are themselves 
modeled by a Dirichlet distribution, so that we finally have
\begin{linenomath*}
\begin{align*}
	\vec{q} &\sim \Dirichlet(\vec{q} \given \frac{\gamma}{K} \cdot \ones{K}) \eqkomma \\
	z_i \given \vec{q} &\sim \Categorical(z_i \given \vec{q}) \eqkomma
\end{align*}
\end{linenomath*}
where $\gamma$ is another concentration parameter, 
controlling the
variability of the mixing coefficients. Note that the indicator variables are still \textit{conditionally} independent given 
the mixing coefficients in this model. More specifically, for a fixed $\vec{q}$, the conditional distribution of a single indicator 
in \cref{eq:condIndicatorFiniteGibbs} and \cref{eq:condIndicatorFiniteCollapsed} takes the following simple form,
\begin{linenomath*}
\begin{equation*}
	\p(z_i=k \given \seq{z}_{\without{i}}, \vec{q}) = q_k \eqpunkt
\end{equation*}
\end{linenomath*}
If 
the value of $\vec{q}$ is unknown, we have two options to include this prior into our model. One is to sample $\vec{q}$ additionally to the remaining variables by drawing 
values from the following conditional distribution during the Gibbs procedure,
\begin{linenomath*}
\begin{align*}
	\p(\vec{q} \given \seq{s}, \seq{a}, \seq{z}, \glob{\Theta}, \alpha) \ &\propto \ \Dirichlet(\vec{q} \given \frac{\gamma}{K}\cdot \ones{K} ) \prod_{i=1}^{|\mathcal{S}|}\Categorical(z_i \given \vec{q}) \\
	& \propto \ \Dirichlet(\vec{q} \given \vec{\zeta} + \frac{\gamma}{K} \cdot \ones{K} ) \eqkomma
\end{align*}
\end{linenomath*}
where $\vec{\zeta}\defeq[\zeta_1\ ,\ \ldots\ ,\ \zeta_K]$, and $\zeta_k$ denotes the number of variables $z_i$ that map to cluster $\mathcal{C}_k$,
\begin{linenomath*}
\begin{equation*}
	\zeta_k = \sum_{i=1}^{|\mathcal{S}|} \indicatorS(z_i=k) \eqpunkt
\end{equation*}
\end{linenomath*}
Alternatively, we can again make use of the conjugacy property to marginalize out the mixing proportions $\vec{q}$ during the inference process, just as we did for the control parameters in previous sections. The result is (additional) collapsing in $\vec{q}$. 
In this case, we simply 
replace the factor $\p(z_i=k \given \seq{z}_{\without{i}})$ 
in the conditional distribution of $z_i$ by 
\begin{linenomath*}
\begin{equation}
\p(z_i=k \given \seq{z}_{\without{i}}, \gamma) \propto (\zeta_k^{(\without{i})}+\frac{\gamma}{K}) \eqkomma
\label{eq:conditionalIndicatorCollapsedMixing}
\end{equation}
\end{linenomath*}
where $\zeta_k^{(\without{i})}$ is defined like $\zeta_k$ but without counting the current value of indicator $z_i$,
\begin{linenomath*}
\begin{equation*}
 \zeta_k^{(\without{i})} \defeq \sum_{\substack{i'=1 \\ i'\neq i}}^{|\mathcal{S}|} \indicatorS(z_i=k) = \zeta_k - \indicatorS(z_i=k) \eqpunkt
\end{equation*}
\end{linenomath*}
A detailed derivation is omitted here but follows the same style as 
for the collapsing in \cref{sec:GibbsCollapsedFinite}. \\[\baselineskip]
\textit{Spatial prior} \nopagebreak[4] \\ 
Both previous prior models assume (conditional) independence of the indicator variables and, hence, make no specific 
assumptions about their 
dependency structure. However, we can also use the prior model to 
promote a certain type of spatial state clustering. 
A reasonable choice is, for instance, to use a model which preferably groups ``similar'' states together (in other words, 
a model which favors clusterings that assign those states the same local control parameter). 
%
%
Similarity of states can be expressed, for example, by a monotonically decreasing decay function $f:[0,\infty)\rightarrow[0,1]$ 
which takes as input the distance between two states. The required pairwise distances can be, in turn, defined via some 
distance metric $\chi:\mathcal{S}\times\mathcal{S}\rightarrow[0,\infty)$.

In fact, apart from the reasons listed in \cref{sec:clusteringApproach}, there is an additional motivation, more intrinsically related to the dynamics of the system, why such a clustering 
can be useful:
given that the transition model of our system admits locally smooth dynamics (which is typically the case for real-world systems), the resulting optimal control policy often turns out 
to be spatially smooth, too \cite{sosic2016}.
More specifically, under an optimal policy, two nearby states are highly likely to experience similar controls; hence, 
it is reasonable to 
assume \textit{a priori} that 
both share the same
local control parameter. 
%
For the policy 
recognition task, it certainly 
makes sense to regularize the 
inference problem by
encoding this particular structure of the solution space into our model.
%
The Potts model \cite{potts1952some}, which is a special case of a Markov random field 
with pairwise clique potentials \cite{koller2009probabilistic}, offers one way to do this,
%
\begin{linenomath*}
\begin{equation*}
	\p_z(\seq{z}) \propto \prod_{i=1}^{|\mathcal{S}|} \exp \Bigg( \frac{\beta}{2} \sum_{\substack{j=1 \\ j\neq i}}^{|\mathcal{S}|} f(d_{i,j}) \delta(z_i, z_j) \Bigg) \eqpunkt
\end{equation*}
\end{linenomath*}
Here, $\delta$ denotes Kronecker's delta, $d_{i,j}\defeq\chi(s_i,s_j),\ i,j\in\lbrace 1,\ldots,|\mathcal{S}|\rbrace$, are the state similarity values, and $\beta\in[0,\infty)$ is the (inverse) temperature of the model 
which controls the strength of the prior. 
%
%
From this
equation, we can easily derive the conditional distribution of a single indicator variable $z_i$ as
\begin{linenomath*}
\begin{equation}
	\p(z_i \given \seq{z}_{\without{i}}) \propto \exp \Bigg( \beta \sum_{\substack{j=1 \\j\neq i}}^{|\mathcal{S}|}  f(d_{i,j}) \delta(z_i, z_j) \Bigg) \eqpunkt 
	\label{eq:PottsConditional}
\end{equation}
\end{linenomath*}
This completes our inference framework for finite 
spaces.


\subsection{Countably infinite and uncountable state spaces}
\label{sec:infiniteStateSpaces}
A major advantage of the clustering approach presented in the last section is that, due to the limited number of 
local policies to be learned from the finite amount of demonstration data, we can now 
apply the same methodology to state spaces of arbitrary size, including countably infinite and uncountable state spaces. 
This extension 
had been 
practically impossible for the static model 
because of the overfitting problem explained in \cref{sec:clusteringApproach}. Nevertheless, there remains a fundamental conceptual problem: a 
direct extension of the model 
to these spaces
would imply that the distribution over possible state partitionings becomes an infinite-dimensional object (\ie, in the case of uncountable state spaces, a distribution over functional mappings from states to 
local controllers), requiring an 
infinite number of indicator variables. Certainly, 
such an object is non-trivial 
to handle computationally. 

However, while the number of latent cluster assignments grows unbounded with the size of the state space, the amount of observed trajectory data 
always remains finite. A 
possible solution to the problem is, therefore, to reformulate the inference task on a reduced 
state space $\widetilde{\mathcal{S}}\defeq\lbrace s_1, s_2, \ldots, s_T\rbrace$ 
containing only states along the observed trajectories. 
Reducing the state space in this way means that we need to 
consider only a finite set of indicator variables $\lbrace z_t \rbrace_{t=1}^T$, one for each expert state $s\in\widetilde{\mathcal{S}}$, which always induces a model of finite size. Assuming that no state is visited twice, we may further use the same index set for both variable \label{foot:indexingP} types.\footnote{\label{foot:indexing} Note that we make this assumption for notational convenience only and that it is not required from a mathematical point of view. Nonetheless, for 
uncountable state spaces the assumption is reasonable 
since the event of reaching the 
same state twice has 
zero probability
for most dynamic models. 
In the general case, however, the indicator variables require their own index set 
to ensure that each system state is associated with exactly one cluster, even when visited multiple times.} In order to 
limit the 
complexity of the dependency structure of the indicator variables for larger data sets, we further let the value of 
indicator $z_t$ depend only on a subset of the remaining variables $\seq{z}_{\without{t}}$ as defined by some neighborhood rule 
$\mathcal{N}$. 
The 
resulting joint distribution is then given as
\begin{linenomath*}
\begin{equation*}
	\p_z(\seq{z} \given \seq{s}) \propto \prod_{t=1}^T\exp \Bigg( \frac{\beta}{2} \sum_{t'\in\mathcal{N}_t} f(d_{t,t'}) \delta(z_t, z_{t'}) \Bigg) \eqkomma
\end{equation*}
\end{linenomath*}
which now implicitly depends on the state sequence $\seq{s}$ through the pairwise distances $d_{t,t'} \defeq \chi(s_t,s_{t'})$, $t,t'\in\{1,\ldots,T\}$ (hence the conditioning on $\seq{s}$).


The use of a finite number of indicator variables along the expert trajectories obviously circumvents the above-mentioned problem of representational complexity. 
Nevertheless, there are some caveats associated with this approach. First of all, using 
a reduced state space model raises the question of 
marginal invariance \cite{blei2011distance}: 
if we added a new trajectory point to the data set, would this change our belief about the expert policy at previously visited states? In particular, how is this different from modeling that new point together with the 
initial ones in the first place? And further, what does such a reduced model imply 
for 
unvisited states? Can we still use it to make predictions about their local policies? These questions are, in fact, important if we plan to use our model to generalize the expert 
demonstrations 
to new situations. For a detailed discussion on this issue, the reader is referred to the supplement. Here, we focus on the 
inferential aspects of the 
problem, which means to identify the 
system parameters at the given trajectory states. 

Another (but related) issue resulting from the reduced modeling approach is that we lose the simple generative interpretation of the process that could be used 
to explain the data generation beforehand. In the 
case of finite state spaces, 
we could think of a trajectory as being constructed by the following, step-wise mechanism: first, the prior $\p_z(\seq{z})$ is used to generate a set of indicator variables for all states. 
Independently, we pick some value for $\alpha$ from $\p_\alpha(\alpha)$ and sample $K$ local control parameters from $\p_\theta(\vec{\theta}_k \given \alpha)$. To finally generate a trajectory, we start with an initial state $s_1$, generated by $\p_1(s_1)$, select a random action $a_1$ from $\pi(a_1 \given s_1, \vec{\theta}_{z_{s_1}})$ and transition to a new state $s_2$ according to $\mathcal{T}(s_2 \given s_1, a_1)$, where we select another action $a_2$, and so on. 
%
Such a directed way of thinking is possible
since the finite model naturally obeys a causal structure where later states depend on earlier ones and the decisions 
made there. 
Furthermore, the cluster assignments and the local controllers could be generated in advance and isolated from each other because they were modeled marginally independent. 

For the reduced state space model, this interpretation no longer applies as 
the model has 
no natural directionality. In fact, 
its variables depend on each other in a cyclic fashion: altering the value of a particular indicator variable (say, the one corresponding to the last trajectory point) will have an effect on the values of 
all remaining indicators due to their spatial relationship encoded by the ``prior distribution'' $\p_z(\seq{z} \given \seq{s})$. Changing the values of the other indicators, however, will influence the actions being played at the respective states which, in turn, alters the probability of ending up with the observed trajectory in the first place and, hence, the position and value of the indicator variable we started with. Explaining the data generation of this model using a simple generative 
process 
is, therefore, 
not possible.

Nevertheless, the individual 
building blocks of our model (that is, the policy, the transition model, \etc) together form a valid 
distribution over the model variables, which can be readily used for parameter inference. For the reasons explained above, it makes sense to define this distribution in the form of a 
discriminative model, ignoring the underlying generative aspects of the process. 
This is sufficient since we can always condition on the observed state sequence $\seq{s}$,
\begin{linenomath*}
\begin{align*}
	\p(\seq{a},\glob{\Theta},\seq{z} &\given \seq{s}, \alpha) = \frac{1}{Z_\seq{s}} \ \p_z(\seq{z} \given \seq{s}) \prod_{k=1}^{K} \p_\theta(\vec{\theta}_k \given \alpha) \ \ldots \\
	&\ldots\ \times\p_1(s_1) \prod_{t=1}^{T-1} \mathcal{T}(s_{t+1} \given s_t, a_t) \pi(a_t \given \vec{\theta}_{z_{s_t}}) \eqpunkt
\end{align*}
\end{linenomath*}
Herein, $Z_\seq{s}$ is a data-dependent normalizing constant. The structure of this distribution is illustrated by the factor graph shown in the supplement (Fig.~S-1), which highlights the circular dependence between the variables. Note that, for any fixed state sequence $\seq{s}$, this distribution indeed encodes the same basic properties as the finite model in \cref{eq:fullClusterModel}.
In particular, the conditional distributions of all remaining variables remain unchanged, which allows us to apply the same inference machinery that we already used 
in the finite case. For a deeper discussion on the difference between the two models, we again point to 
	the supplement.

\section{Nonparametric Policy Recognition}
\label{sec:nonparametric}
In the last section, we presented a probabilistic policy recognition framework 
for modeling the expert behavior using a finite mixture of $K$ local policies.
Basically, there are two situations when such a model 
is useful:
\begin{itemize}
	\item either, we know 
	the true number of 
	expert policies, 
	\item or, irrespective of the true behavioral 
	complexity, we 
	want to find an approximate system description 
	in terms of at most $K$ distinct control situations \cite{waltz1965heuristic} (\cf\ finite state controllers \cite{meuleau1999learning}). 
\end{itemize}
In all other cases, we are faced with the non-trivial problem of 
choosing
$K$. 
In fact, the choice of $K$ should not just be considered
a mathematical necessity to perform inference in our model. 
By selecting a certain value for $K$ we
can, of course, directly control the complexity class of 
potentially inferred expert 
controllers. However, from a system identification point of view, it is more reasonable to infer the 
required granularity of the state partitioning from the observed 
expert behavior itself, instead of enforcing a particular model complexity. 
This way, we can gain valuable information
about the underlying control structure and 
state representation 
used by the expert, 
which offers a possibility to 
learn a state partitioning of task-appropriate complexity
directly from the demonstration data. Hence, 
the problem of selecting 
the right model structure should be considered as part of the 
inference 
problem
itself. 

From a 
statistical modeling perspective, there are two 
common ways to approach 
this problem. One is to make use of model selection techniques in order to determine the most 
parsimonious 
model that is 
in agreement with the observed data.
However, choosing a particular 
model complexity 
still means that we consider only one possible 
explanation for 
the data, although other explanations might be likewise plausible. 
For many inference tasks, including this one, the more elegant approach is to 
keep the complexity flexible and, hence, adaptable to the data. Mathematically, this can be achieved by 
assuming a 
potentially
infinite set of 
model parameters (in our case controllers)
from which we activate only a finite subset
to explain the particular data set at hand. This alternative way of thinking opens the door to the rich class of nonparametric models, which provide an integrated 
framework to formulate the inference problem over both model parameters and model complexity as a joint learning problem. 


\subsection{A Dirichlet process mixture model}
The classical way to nonparametric clustering is to use a Dirichlet process mixture model (DPMM) \cite{neal2000markov}. These models can be obtained by starting from a finite mixture model and letting the number of mixture components (\ie\ the number of local controllers) approach infinity. In our case, we start with the 
clustering model from \cref{sec:clusteringApproach}, using a mixing prior 
over indicator variables, 
\begin{linenomath*}
\begin{equation}
\hspace{-1.5cm}
\begin{minipage}{0.4\columnwidth}
\begin{tabular}{rcl}
 	$\vec{q}$ \hspace{-4mm} & $\sim$ \hspace{-4mm} & $\Dirichlet(\vec{q} \given \frac{\gamma}{K} \cdot \ones{K})$	
	\\[2mm] $\vec{\theta}_k$ \hspace{-4mm} & $\sim$ \hspace{-4mm} & $\Dirichlet(\vec{\theta}_k \given \alpha \cdot \ones{|\mathcal{A}|})$ 	     
	\\[2mm] $s_1$ \hspace{-4mm} & $\sim$ \hspace{-4mm} & $\p_1(s_1)$ 	
\end{tabular}
\end{minipage}
\hspace{0.4cm}
\begin{minipage}{0.4\columnwidth}
\begin{tabular}{rcl}
 	$z_i \given \vec{q}$ \hspace{-4mm} & $\sim$ \hspace{-4mm} &  $\Categorical(z_i \given \vec{q})$
	\\[2mm] $a_t \given s_t, \glob{\Theta}, \seq{z}$ \hspace{-4mm}  & $\sim$ \hspace{-4mm} & $\pi(a_t \given \vec{\theta}_{z_{s_t}})$
	\\[2mm] $s_{t+1} \given s_t, a_t$ \hspace{-4mm} & $\sim$ \hspace{-4mm} & $\mathcal{T}(s_{t+1} \given s_t, a_t)$ \eqpunkt
\end{tabular}
\end{minipage}
\label{eq:FMMtoDPMM}
\end{equation}
\end{linenomath*}
From 
these equations, we arrive at the corresponding nonparametric 
model 
as $K$ goes to infinity. 
For the theoretical foundations of this limit, the reader is referred to the more general literature on Dirichlet processes, such as \cite{ferguson1973bayesian,neal2000markov}. In this paper, we restrict ourselves to providing the 
resulting sampling mechanisms for the policy recognition problem. 

In 
a DPMM, the mixing proportions $\vec{q}$ of 
the local 
parameters are marginalized out (that is, we use a collapsed sampler). 
The resulting distribution over partitionings is described by a Chinese restaurant process (CRP) \cite{aldous1985exchangeability} 
which can be derived, for instance, 
by considering the limit $K\rightarrow\infty$ of the mixing process 
induced by the Gibbs update in \cref{eq:conditionalIndicatorCollapsedMixing},
\begin{linenomath*}
\begin{equation}
	\p(z_i=k \given \seq{z}_{\without{i}}, \gamma) \propto 
	\begin{cases}
		\zeta_k^{(\without{i})}  & \text{if } k\in\lbrace 1,\ldots,K^* \rbrace \eqkomma\\
		\gamma & \text{if } k=K^*+1 \eqpunkt
	\end{cases}
	\label{eq:CRPprior}
\end{equation}
\end{linenomath*}
Here, $K^*$ denotes the number of distinct entries in $\seq{z}_{\without{i}}$ which are 
represented by the numerical 
values $\lbrace 1, \ldots, K^*\rbrace$. 
In this model, a 
state joins 
an existing cluster (\ie\ a group of states whose indicators have the same value) with probability proportional to the number of states already contained in that 
cluster. Alternatively, 
it may create a new cluster with probability proportional to $\gamma$. 

From the model equations 
\eqref{eq:FMMtoDPMM} it is evident that, given a particular setting of indicators, 
the conditional distributions of all other variable types remain unchanged. 
Effectively, we only replaced the prior model $p_z(\seq{z})$ by the CRP. Hence, we can apply the same Gibbs updates for the actions and controllers as before 
and need to rederive only the conditional distributions of the indicator variables 
under consideration of the above defined process. According to \cref{eq:CRPprior}, we herein need to distinguish whether an indicator variable takes a value already occupied by other indicators (\ie\ it joins an existing cluster) or it is assigned a new value (\ie\ it creates a new cluster). 
Let $\lbrace \vec{\theta}_k \rbrace_{k=1}^{K^*}$ denote the set of control parameters associated with 
$\seq{z}_{\without{i}}$. In the first case ($k\in\lbrace 1, \ldots, {K^*}\rbrace$), we can then write
\begin{linenomath*}
\begin{align*}
	&\p(z_i=k \given \seq{z}_{\without{i}}, \seq{s}, \seq{a}, \lbrace \vec{\theta}_{k'} \rbrace_{k'=1}^{K^*}, \alpha, \gamma) \\
	&= \p(z_i=k \given \seq{z}_{\without{i}}, \lbrace a_t \rbrace_{t:s_t=i}, \vec{\theta}_{k}, \alpha, \gamma) \\
	&\propto \p(z_i=k \given \seq{z}_{\without{i}}, \vec{\theta}_{k}, \alpha, \gamma) \p(\lbrace a_t \rbrace_{t:s_t=i} \given z_i=k, \seq{z}_{\without{i}}, \vec{\theta}_{k}, \alpha, \gamma) \\
	&\propto \p(z_i=k \given \seq{z}_{\without{i}}, \gamma) \p(\lbrace a_t \rbrace_{t:s_t=i} \given \vec{\theta}_{k}) \\
	&\propto \zeta_k^{(\without{i})} \cdot \prod_{t:s_t=i}\pi(a_t \given \vec{\theta}_{k}) \eqpunkt
\end{align*}
\end{linenomath*}
In the second case ($k={K^*}+1$), we instead obtain
\begin{linenomath*}
\begin{align*}
	\p&(z_i={K^*}+1 \given \seq{z}_{\without{i}}, \seq{s}, \seq{a}, \lbrace \vec{\theta}_k \rbrace_{k=1}^{K^*}, \alpha, \gamma) \\
	&= \p(z_i={K^*}+1 \given \seq{z}_{\without{i}}, \lbrace a_t \rbrace_{t:s_t=i}, \alpha, \gamma) \\
	&\propto \p(z_i={K^*}+1 \given \seq{z}_{\without{i}}, \alpha, \gamma) \ \ldots \\
	&\phantom{\propto} \ \ \ldots \ \times \ \p(\lbrace a_t \rbrace_{t:s_t=i} \given z_i={K^*}+1, \seq{z}_{\without{i}}, \alpha, \gamma)  \\
	&\propto \p(z_i={K^*}+1 \given \seq{z}_{\without{i}}, \gamma) \p(\lbrace a_t \rbrace_{t:s_t=i} \given z_i={K^*}+1,\alpha) \\
	&\propto \gamma \cdot \integralR{\p(\lbrace a_t \rbrace_{t:s_t=i} \given \vec{\theta}_{K^*+1}) \p_\theta(\vec{\theta}_{K^*+1} \given \alpha)}{\vec{\theta}_{K^*+1}}{\Delta} \\
	&\propto \gamma \cdot \integralR{\prod_{t:s_t=i}\pi(a_t \given \vec{\theta}_{K^*+1}) \p_\theta(\vec{\theta}_{K^*+1} \given \alpha)}{\vec{\theta}_{K^*+1}}{\Delta} \\
	&\propto \gamma \cdot \DirMult(\vec{\phi}_i \given \alpha) \eqpunkt
\end{align*}
\end{linenomath*}
If a new cluster is created, we further need to initialize the corresponding control parameter $\vec{\theta}_{K^*+1}$ by performing the respective Gibbs update, \ie\ by sampling from
\begin{linenomath*}
\begin{align*}
	\p(\vec{\theta}_{K^*+1} &\given \seq{z}, \seq{s}, \seq{a}, \lbrace \vec{\theta}_k \rbrace_{k=1}^{K^*}, \alpha, \gamma)  \\
	&= \p(\vec{\theta}_{K^*+1} \given \lbrace a_t \rbrace_{t:z_{s_t}={K^*+1}}, \alpha) \\
	&\propto \p_\theta(\vec{\theta}_{K^*+1} \given \alpha) \p(\lbrace a_t \rbrace_{t:z_{s_t}={K^*+1}} \given \vec{\theta}_{K^*+1})  \\
	&\propto \p_\theta(\vec{\theta}_{K^*+1} \given \alpha) \hspace{3mm} \cdot \hspace{-3mm} \prod_{t:z_{s_t}={K^*+1}} \pi( a_t \given \vec{\theta}_{K^*+1}) \\
	&\propto \Dirichlet(\vec{\theta}_{K^*+1} \given \vec{\xi}_{K^*+1} + \alpha \cdot \ones{|\mathcal{A}|}) \eqpunkt
\end{align*}
\end{linenomath*}
Should a cluster get unoccupied during the sampling process, the corresponding control 
parameter may be removed from the stored parameter set $\lbrace \vec{\theta}_k \rbrace$ 
and the index set for $k$ 
needs to be updated accordingly. Note that 
this sampling mechanism is a specific instance of Algorithm~2 described in \cite{neal2000markov}. A collapsed variant can be derived in a similar fashion. 

\subsection{Policy recognition using the distance-dependent Chinese restaurant process}
In the previous section, we have seen that the DPMM can be 
derived as the nonparametric limit model of a finite mixture 
using a set of latent mixing proportions $\vec{q}$ for the clusters. Although the DPMM allows us to keep the number of active controllers flexible and, hence, adaptable to 
the complexity of the demonstration data, 
the CRP as the underlying clustering mechanism does not capture any spatial dependencies between 
the indicator variables. 
In fact, in the CRP, the indicators $\{z_i\}$ are coupled only via 
their relative frequencies (see \cref{eq:CRPprior}) but not through their individual locations in space, 
resulting in an exchangeable collection of random variables \cite{aldous1985exchangeability}. 
In fact, one could argue that 
the spatial structure of the 
clustering problem is \textit{a~priori} ignored.

The fact that 
DPMMs 
are nevertheless used for spatial clustering tasks can be explained by the particular form of data likelihood models that are typically used for the mixture components. In a Gaussian mixture model \cite{Rasmussen00theinfinite}, for instance, 
the spatial clusters emerge due to the unimodal nature of the mixture components, 
which 
encodes the locality property of the model 
needed 
to obtain a meaningful spatial clustering of the data. 
For the policy recognition problem, however, the DPMM is not able to exploit any spatial information 
via the data likelihood
since the clustering of states is performed at the level of the inferred action information (see \cref{eq:condIndicatorFiniteGibbs}) 
and not on the state sequence itself. Consequently, we cannot expect to obtain a smooth clustering of the system state space, 
especially when the expert policies are overlapping (\ie\ when they share one or more common actions) so that 
the action information alone is not sufficient to discriminate between policies. For uncountable state spaces, this problem is further complicated by the fact that we observe \textit{at most} one expert state transition per system state. Here, the spatial 
context of the data is the only information which can resolve this ambiguity.

In order to facilitate a spatially smooth clustering, we therefore need to 
consider  
non-exchangeable distributions over partitionings. More specifically, 
we 
need to design our model in such a way that, 
whenever a state $s$ is ``close'' to some other state $s'$ and assigned to some cluster $\mathcal{C}_k$, then, \textit{a priori}, $s'$ should 
belong to the same cluster $\mathcal{C}_k$ with high probability. In that sense, we are looking for the nonparametric counterpart of the Potts model. 
\begin{figure}
	\centering
	\includegraphics[width=6cm]{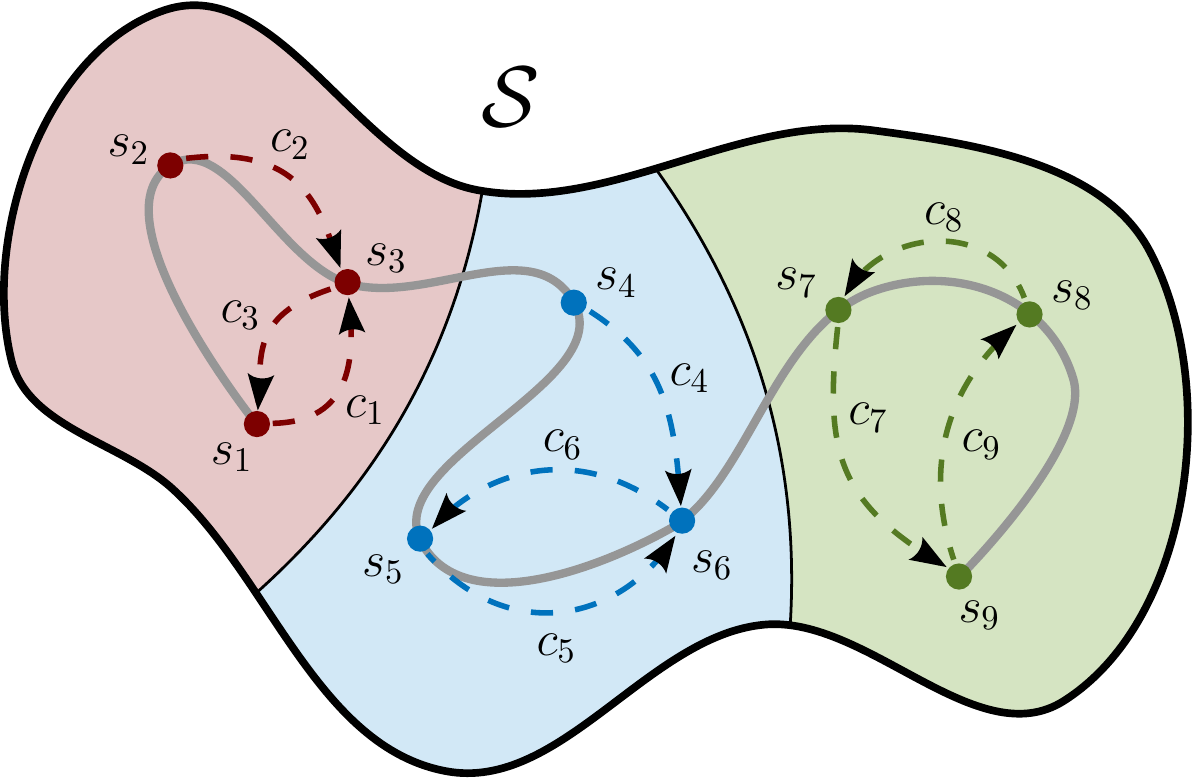}
	\caption{Schematic illustration of the ddCRP-based clustering applied to the reduced state space model in \cref{sec:infiniteStateSpaces}. Each trajectory state is connected to some other state of the sequence. The connected components of the resulting graph implicitly define the state clustering. Coloring of the background illustrates the spatial cluster extrapolation (see Section~A in the supplement). Note that the underlying decision-making process is assumed to be discrete in time; the continuous gray line shown in the figure is only to indicate the temporal ordering of the trajectory states.}
	\label{fig:ddCRPtrajectory}
\end{figure}
One model with such properties is the distance-dependent Chinese restaurant process (ddCRP) \cite{blei2011distance}.\footnote{Note that the authors of \cite{blei2011distance} avoid calling this model nonparametric since it cannot 
be cast as a mixture model originating from a random measure. However, we stick to this term 
in order to make a clear distinction to the parametric models in \cref{sec:parametricPolicyRecognition}, and to highlight the fact that there is no parameter $K$ 
determining the number of controllers.} As opposed to the traditional CRP, the ddCRP explicitly takes into account the spatial 
structure of 
the data. This is done in the form of pairwise distances 
between 
states, 
which can be obtained, for instance, by defining an appropriate distance metric on the state space 
(see \cref{sec:priorModels}). 
Instead of assigning states to clusters as done by the CRP, the ddCRP assigns states 
to other states according to their pairwise distances. More specifically, the probability that state $i$ gets assigned to state $j$ is defined as
\begin{linenomath*}
\begin{equation}
	\p(c_i=j \given \mat{D}, \nu) \ \propto \ \begin{cases} \nu & \text{if } i=j \eqkomma \\ f(d_{i,j}) & \text{otherwise}\eqkomma \end{cases}
	\label{eq:ddcrpPrior}
\end{equation}
\end{linenomath*}
	where $\nu\in[0,\infty)$ is called the self-link parameter of the process, $\mat{D}$ denotes the collection of all pairwise state distances, and $c_i$ is the ``to-state'' assignment of state $i$, which can be thought of as a directed edge on the graph defined on the set of all states 
(see \cref{fig:ddCRPtrajectory}). Accordingly, 
$i$ and $j$ in \cref{eq:ddcrpPrior} can take values 
$\lbrace1,\ldots,|\mathcal{S}|\rbrace$ for the finite state space model and 
$\lbrace1,\ldots,T\rbrace$ for our reduced state space model. 
The 
	state clustering is then obtained as a byproduct of this mapping 
via the connected components of the resulting graph 
(see \cref{fig:ddCRPtrajectory} again). 


Replacing the CRP by the ddCRP and following 
the same line of argument as in \cite{blei2011distance}, we obtain the required 
conditional distribution 
of the state assignment $c_i$ as
\begin{linenomath*}
\begin{align*}
	\p(c_i\!=\!j \!\given\! \seq{c}_{\without{i}}, \hspace{-0.15ex}\seq{s}, \hspace{-0.15ex}\seq{a}, \hspace{-0.15ex}\alpha, \hspace{-0.15ex}\mat{D}, \hspace{-0.1ex}\nu) \!\propto\!\!
	\begin{cases}
		\nu &  \hspace{-1ex} j=i \eqkomma \\
		f(d_{i,j}) & \hspace{-1ex} \text{no clusters merged} \eqkomma \\
		f(d_{i,j})\!\cdot\!\mathcal{L}  & \hspace{-1ex} \text{$\mathcal{C}_{z_i}$ and $\mathcal{C}_{z_j}$ merged} \eqkomma
	\end{cases}
\end{align*} 
\end{linenomath*}
where we use the shorthand notation
\begin{linenomath*}
\begin{equation*}
	\mathcal{L} = \frac{\DirMult(\vec{\xi}_{z_i} + \vec{\xi}_{z_j} \given \alpha)}{\DirMult(\vec{\xi}_{z_i} \given \alpha)\DirMult(\vec{\xi}_{z_j} \given \alpha)}
\end{equation*}
\end{linenomath*}
for the data likelihood term. 
The $\xi_{k,j}$'s are defined as in \cref{eq:definitionPolicyActionCounts} 
but are based on the clustering which arises when we ignore the current link $c_i$. 
The resulting Gibbs sampler is a collapsed one as the local control parameters are 
necessarily marginalized out during the inference process. 

\section{Simulation Results}
\label{sec:results}
In this section, we present simulation results for two types of system dynamics. 
As a proof-of-concept, we 
first investigate the case of 
uncountable state spaces which we consider 
the more challenging setting for reasons explained earlier. 
To compare our framework with existing methods, we further provide simulation results for the standard grid world benchmark (see \eg\ \cite{sosic2016,michini2012bayesian,abbeel2004apprenticeship}). It should be pointed out, however, that establishing a fair comparison between LfD models is generally 
difficult due to their different working principles (\eg\ reward prediction vs.\ action prediction), objectives (system identification vs.\ optimal control), requirements (\eg\ MDP solver, knowledge of the expert's discount factor, 
countable vs.\ uncountable state space), and assumptions (\eg\ deterministic vs.\ stochastic expert behavior).  
Accordingly, our goal is rather to demonstrate the prediction abilities of the considered models than 
to push the models to their individual limits. Therefore, and to reduce the overall computational load, we tuned most model hyper-parameters by hand. 
Our code is available at \url{https://github.com/AdrianSosic/BayesianPolicyRecognition}.

\subsection{Example 1: uncountable state space}
\label{sec:continuousExample}
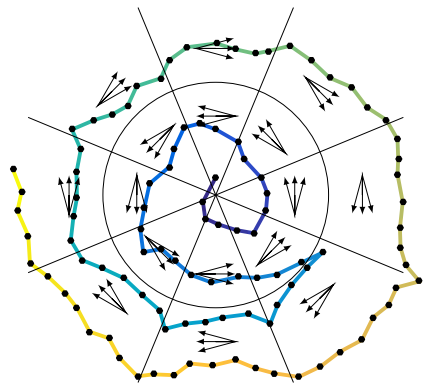
\begin{figure}
\centering
	\scalebox{0.75}{
		\begin{tikzpicture}
			\node {\includegraphics[scale=0.96,trim=30 30 30 30,clip]{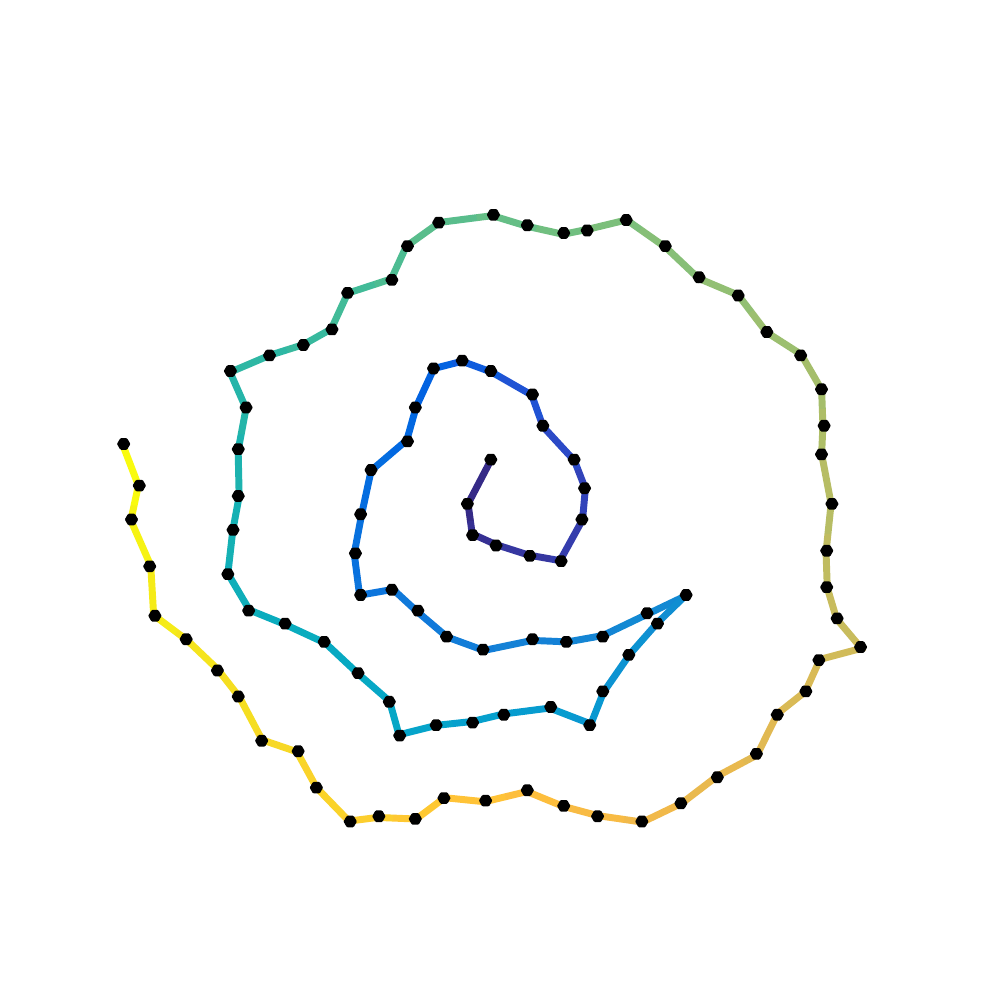}};
		
			\draw (0,0) circle (2cm);
			\draw (22.5:\ax) -- (202.5:\ax);
			\draw (67.5:\ax) -- (247.5:\ax);
			\draw (-22.5:\ax) -- (-202.5:\ax);
		    \draw (-67.5:\ax) -- (-247.5:\ax);
		    
		    
		    \draw [-latex] (315:\dist) --++ (225:\off) --+ (60:\lw);
		    \draw [-latex] (315:\dist)  --++ (225:\off) --+ (45:\lw);
		    \draw [-latex] (315:\dist)  --++ (225:\off) --+ (30:\lw);
		    
		    \draw [-latex] (270:\dist) --++ (180:\off) --+ (15:\lw);
		    \draw [-latex] (270:\dist)  --++ (180:\off) --+ (0:\lw);
		    \draw [-latex] (270:\dist)  --++ (180:\off) --+ (345:\lw);
		    
		    \draw [-latex] (225:\dist) --++ (135:\off) --+ (330:\lw);
		    \draw [-latex] (225:\dist)  --++ (135:\off) --+ (315:\lw);
		    \draw [-latex] (225:\dist)  --++ (135:\off) --+ (300:\lw);
		    
		    \draw [-latex] (180:\dist) --++ (90:\off) --+ (285:\lw);
		    \draw [-latex] (180:\dist)  --++ (90:\off) --+ (270:\lw);
		    \draw [-latex] (180:\dist)  --++ (90:\off) --+ (255:\lw);
		    
		    \draw [-latex] (135:\dist) --++ (45:\off) --+ (240:\lw);
		    \draw [-latex] (135:\dist)  --++ (45:\off) --+ (225:\lw);
		    \draw [-latex] (135:\dist)  --++ (45:\off) --+ (210:\lw);
		    
		    \draw [-latex] (90:\dist) --++ (0:\off) --+ (195:\lw);
		    \draw [-latex] (90:\dist)  --++ (0:\off) --+ (180:\lw);
		    \draw [-latex] (90:\dist)  --++ (0:\off) --+ (165:\lw);

		    \draw [-latex] (45:\dist) --++ (-45:\off) --+ (150:\lw);
		    \draw [-latex] (45:\dist)  --++ (-45:\off) --+ (135:\lw);
		    \draw [-latex] (45:\dist)  --++ (-45:\off) --+ (120:\lw);
		    
		    \draw [-latex] (0:\dist) --++ (-90:\off) --+ (105:\lw);
		    \draw [-latex] (0:\dist)  --++ (-90:\off) --+ (90:\lw);
		    \draw [-latex] (0:\dist)  --++ (-90:\off) --+ (75:\lw);
		    
		    
		    \draw [-latex] (315:-\distl) --++ (225:\off) --+ (60:\lw);
		    \draw [-latex] (315:-\distl)  --++ (225:\off) --+ (45:\lw);
		    \draw [-latex] (315:-\distl)  --++ (225:\off) --+ (30:\lw);
		    
		    \draw [-latex] (270:-\distl) --++ (180:\off) --+ (15:\lw);
		    \draw [-latex] (270:-\distl)  --++ (180:\off) --+ (0:\lw);
		    \draw [-latex] (270:-\distl)  --++ (180:\off) --+ (345:\lw);
		    
		    \draw [-latex] (225:-\distl) --++ (135:\off) --+ (330:\lw);
		    \draw [-latex] (225:-\distl)  --++ (135:\off) --+ (315:\lw);
		    \draw [-latex] (225:-\distl)  --++ (135:\off) --+ (300:\lw);
		    
		    \draw [-latex] (180:-\distl) --++ (90:\off) --+ (285:\lw);
		    \draw [-latex] (180:-\distl) --++ (90:\off) --+ (270:\lw);
		    \draw [-latex] (180:-\distl)  --++ (90:\off) --+ (255:\lw);
		    
		    \draw [-latex] (135:-\distl) --++ (45:\off) --+ (240:\lw);
		    \draw [-latex] (135:-\distl)  --++ (45:\off) --+ (225:\lw);
		    \draw [-latex] (135:-\distl)  --++ (45:\off) --+ (210:\lw);
		    
		    \draw [-latex] (90:-\distl) --++ (0:\off) --+ (195:\lw);
		    \draw [-latex] (90:-\distl)  --++ (0:\off) --+ (180:\lw);
		    \draw [-latex] (90:-\distl)  --++ (0:\off) --+ (165:\lw);

		    \draw [-latex] (45:-\distl) --++ (-45:\off) --+ (150:\lw);
		    \draw [-latex] (45:-\distl)  --++ (-45:\off) --+ (135:\lw);
		    \draw [-latex] (45:-\distl)  --++ (-45:\off) --+ (120:\lw);
		    
		    \draw [-latex] (0:-\distl) --++ (-90:\off) --+ (105:\lw);
		    \draw [-latex] (0:-\distl)  --++ (-90:\off) --+ (90:\lw);
		    \draw [-latex] (0:-\distl)  --++ (-90:\off) --+ (75:\lw);
		\end{tikzpicture}
	}
\vspace{-2ex}
\caption{Schematic illustration of the expert policy used in \cref{sec:continuousExample}, which applies eight local controllers to sixteen distinct regions. A sample trajectory is shown in color.}
\label{subfig:schematicOptimalPolicy}
\end{figure}

As 
an illustrative example, we consider 
a 
dynamical system which describes 
the circular motion of an agent on a 
two-dimensional state space. 
%
The actions of the agent correspond to  
24 directions that divide the space of possible angles $[0,2\pi)$ into equally-sized intervals. More specifically, action~$j$ corresponds to the angle $(j-1)\frac{2\pi}{24}$.
The transition model of the system is defined as follows: for each selected action, the agent first makes a step of length $\mu=1$ in the 
intended direction. 
The so-obtained position is then distorted by additive zero-mean isotropic Gaussian noise of variance~$\sigma^2$. This defines our transition kernel as 
\begin{linenomath*}
\begin{equation}
	\mathcal{T}(s_{t+1} \given s_t, a_t=j) = \Gaussian(s_{t+1} \given s_t + \mu \cdot \mathbf{e}_j, \ \sigma^2 \idMat) \eqkomma
	\label{eq:transitionKernel}
\end{equation}
\end{linenomath*}
where $s_t,s_{t+1}\in\reals^2$, $\mathbf{e}_j$ denotes the two-dimensional unit vector pointing in the direction 
of action $j$, and $\idMat$ is the two-dimensional identity matrix.
The overall goal of our agent is to describe a circular motion around the origin in the best possible manner allowed by the 
available actions. However, due to 
limited sensory information, the agent is not able to observe its exact position on the plane 
but can only distinguish between certain regions of the state space, as illustrated by \cref{subfig:schematicOptimalPolicy}. Also, the agent is unsure about the optimal control strategy, \ie\ it 
does not always 
make 
optimal decisions but selects its actions uniformly at random from a subset of actions, consisting of the optimal one and the two actions pointing to neighboring directions (see \cref{subfig:schematicOptimalPolicy} again). 
To increase the difficulty of the prediction task, we further let the agent change the direction of travel whenever 
the critical distance of $r=5$ to the origin is exceeded.

Having defined the expert behavior, we generate 10 sample trajectories 
of length $T=100$. Herein, we assume a motion noise level of $\sigma=0.2$ and initialize the agent's position uniformly at random on the unit circle. An example trajectory is shown in \cref{subfig:schematicOptimalPolicy}. The obtained trajectory data is fed into the 
presented inference algorithms to approximate the posterior distribution over 
expert controllers, and the whole experiment is repeated in 100 Monte Carlo runs. 

For the spatial models, we use the Euclidean metric to compute the pairwise distances between states,
\begin{linenomath*}
\begin{equation}
	\chi(s,s') = ||s-s'||_2 \eqpunkt
	\label{eq:euclideanMetric}
\end{equation}
\end{linenomath*}
The corresponding similarity values are calculated using a Gaussian-shaped kernel. More specifically,
\begin{linenomath*}
\begin{equation*}
	f_\text{Potts}(d) = \exp\left(-\frac{d^2}{\sigma_f^2}\right)
\end{equation*}
\end{linenomath*}
for the Potts model and
\begin{linenomath*}
\begin{equation*}
	f_\text{ddCRP}(d) = (1-\epsilon)f_\text{Potts}(d) + \epsilon
\end{equation*}
\end{linenomath*}
for the ddCRP model, with $\sigma_f=1$ and a constant offset of $\epsilon=0.01$ which ensures that 
states with large distances can still join the same cluster. 
For the Potts model, we further use a neighborhood structure 
containing the eight closest trajectory points of a state. This way, we ensure that, in principle, each local expert policy may occur at least once in the neighborhood of a state. The concentration parameter for the local controls is set to $\alpha=1$, corresponding to a uniform prior belief over local policies.

A major drawback of the Potts model is that posterior inference about the temperature parameter $\beta$ 
is complicated 
due to the nonlinear effect of the parameter on the normalization of the model. Therefore, we manually selected a temperature of $\beta=1.6$ based on a minimization of the average policy prediction error (discussed below) via parameter sweeping. As opposed to this, we extend the inference problem for the ddCRP to the self-link parameter~$\nu$ 
as suggested in \cite{blei2011distance}. For this, we use an exponential prior,
\begin{linenomath*}
\begin{equation*}
\p_\nu(\nu) = \Exponential(\nu \given \lambda) \eqkomma
\end{equation*}
\end{linenomath*}
with rate parameter $\lambda=0.1$, and applied the independence Metropolis-Hastings algorithm \cite{chib1995understanding} 
using $\p_\nu(\nu)$ as proposal distribution 
with an initial value of $\nu=1$. In all our simulations, 
the sampler quickly converged to its stationary distribution, 
yielding posterior values for $\nu$ with a mean of $0.024$ and a standard deviation of $0.023$.

\begin{figure}
	\hspace{-2mm} \includegraphics[scale=0.5]{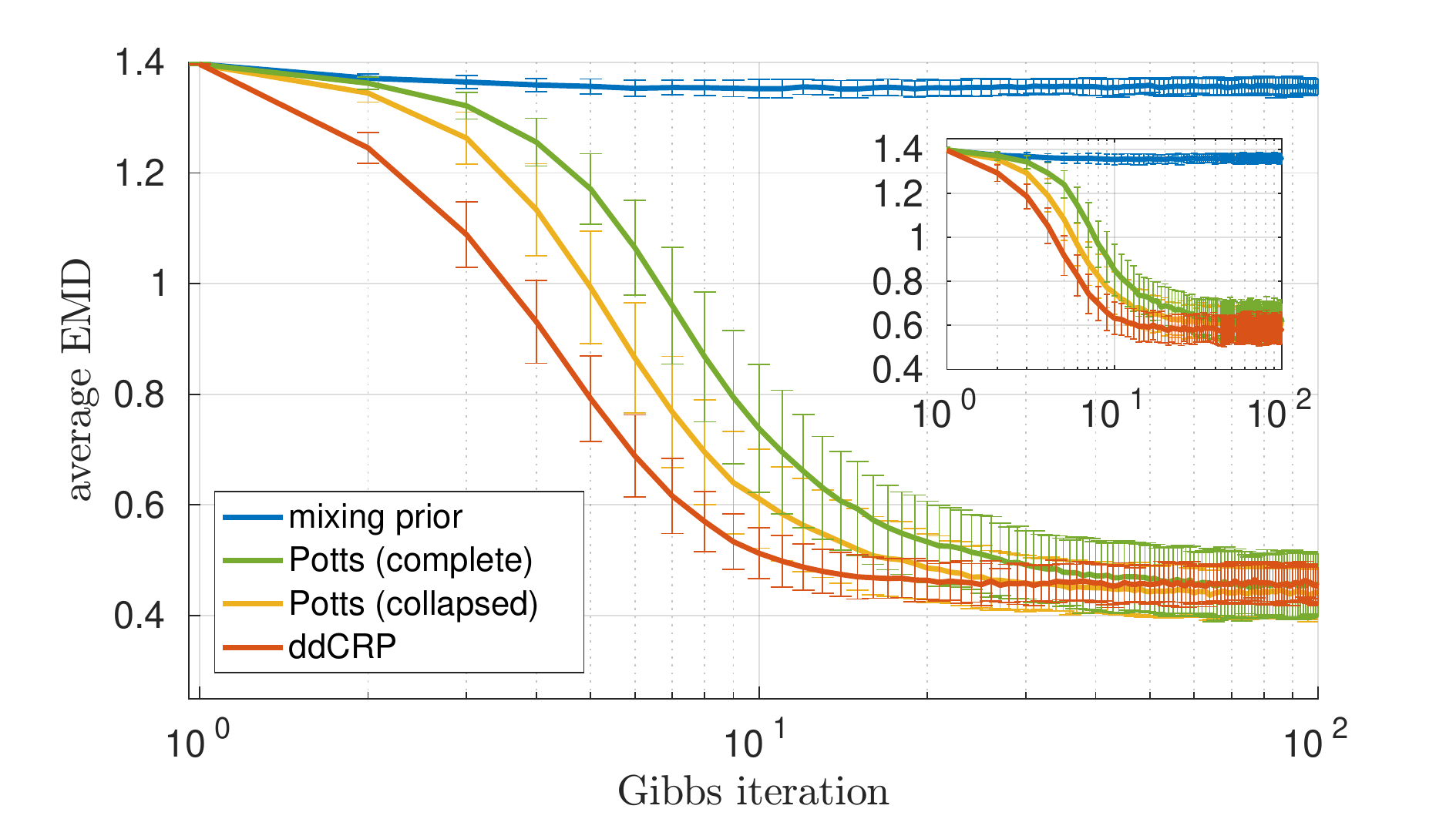}
	\caption{Average policy prediction error at the simulated trajectory states (main figure) and at non-trajectory states (inset).  
Shown are the empirical mean values and standard deviations, 
estimated from 100 Monte Carlo runs.} 
	\label{fig:learningProgress}
\end{figure}

To locally compare the predicted policy with the ground truth at a given state, we compute their earth mover's distance (EMD) \cite{rubner1998metric} with a ground distance metric measuring the absolute angular difference between 
the involved actions. 
%
%
To track the 
learning progress of the 
algorithms, we calculate the average EMD over all states of the given trajectory set at each Gibbs iteration. Herein, the local policy predictions are computed from the single Gibbs sample of the respective iteration, consisting of all sampled actions, indicators and -- in case of non-collapsed sampling -- the local control parameters. 
The resulting mean EMDs and standard deviations are depicted in \cref{fig:learningProgress}. 
The inset 
further shows the average EMD computed at non-trajectory states 
which are sampled on a regular grid (depicted in the supplement), reflecting the quality of the resulting spatial prediction. 

As expected, the finite mixture model (using 
the true number of local policies, a collapsed mixing prior, and $\gamma=1$) is not able to learn a reasonable policy representation from the expert demonstrations since it does not 
explore the spatial structure of the data. In fact, the resulting prediction error shows only a slight improvement as compared to 
an untrained model. 
In contrast to this, all spatial models capture 
the expert behavior reasonably well. In agreement with our reasoning in \cref{sec:GibbsCollapsedFinite}, we observe that the collapsed Potts 
model 
mixes significantly faster and has a smaller prediction variance than the non-collapsed version. However, the ddCRP model gives the best result, both in terms of mixing speed (see \cite{blei2011distance} for an explanation of this phenomenon) and model accuracy. 
Interestingly, this is despite the fact that the ddCRP model additionally needs to infer the number of local controllers necessary to reproduce the expert behavior. 
The corresponding 
posterior distribution, 
which shows a pronounced peak at the true number, is depicted in the supplement. 
There, we also provide additional simulation results which give insights into the learned state partitioning and the resulting spatial policy prediction error. The results reveal that all expert motion patterns can be 
identified by our algorithm. 
%

\subsection{Example 2: finite state space}
\label{sec:discreteSetting}

\begin{figure*}
	\centering
	\begin{subfigure}{6.0cm}
		\centering
		\includegraphics[scale=0.5]{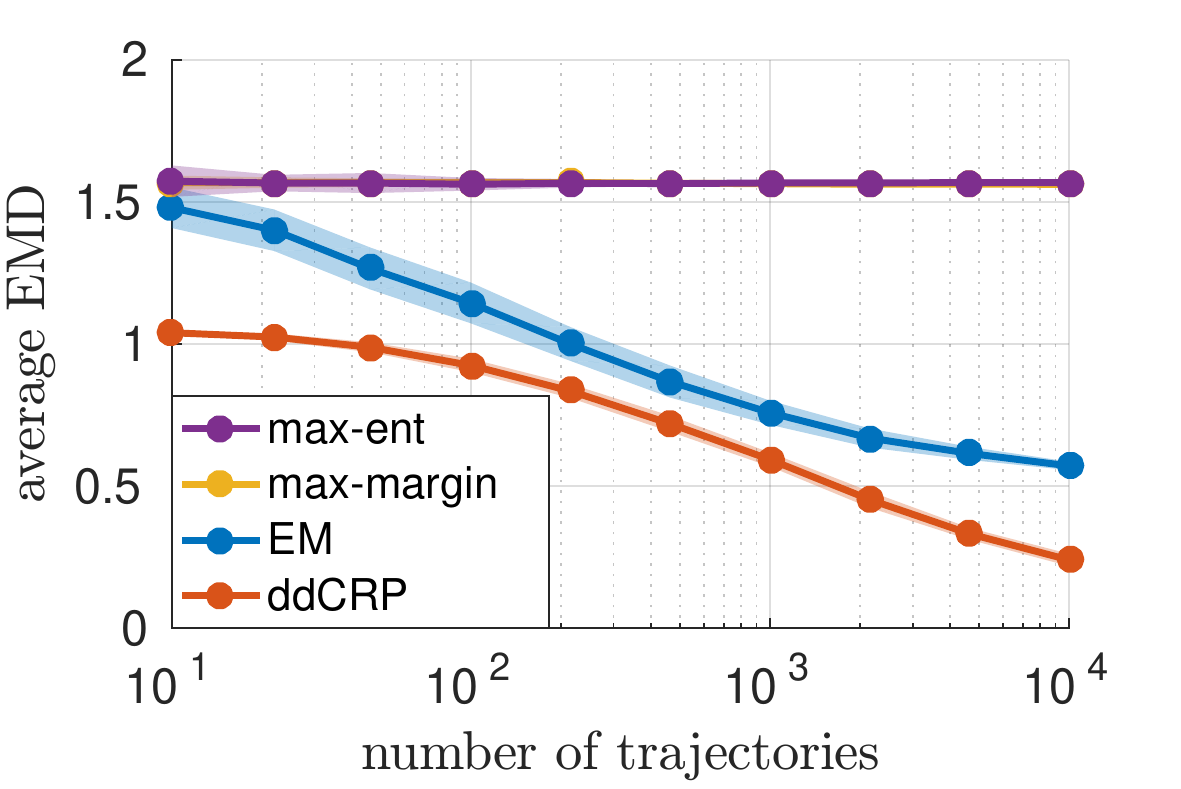}
		\caption{learning curves (circular policy)}
		\label{subfig:circularGridWorld}
	\end{subfigure}
	\hspace{-0.1cm}
	\begin{subfigure}{6cm}
		\centering
		\includegraphics[scale=0.5]{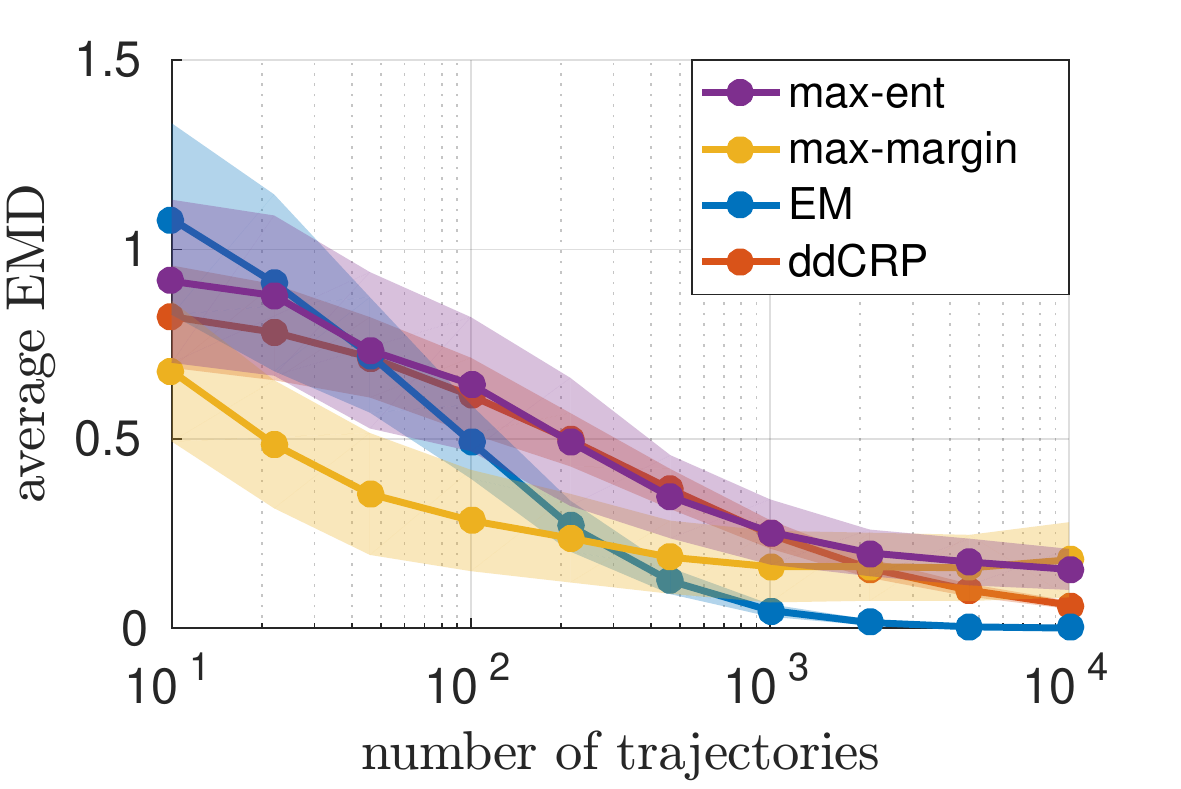}
		\caption{learning curves (MDP policy)}
		\label{subfig:MDPGridWorld}
	\end{subfigure}
	\hspace{-0.1cm}
	\begin{subfigure}{6cm}
		\centering
		\includegraphics[scale=0.5]{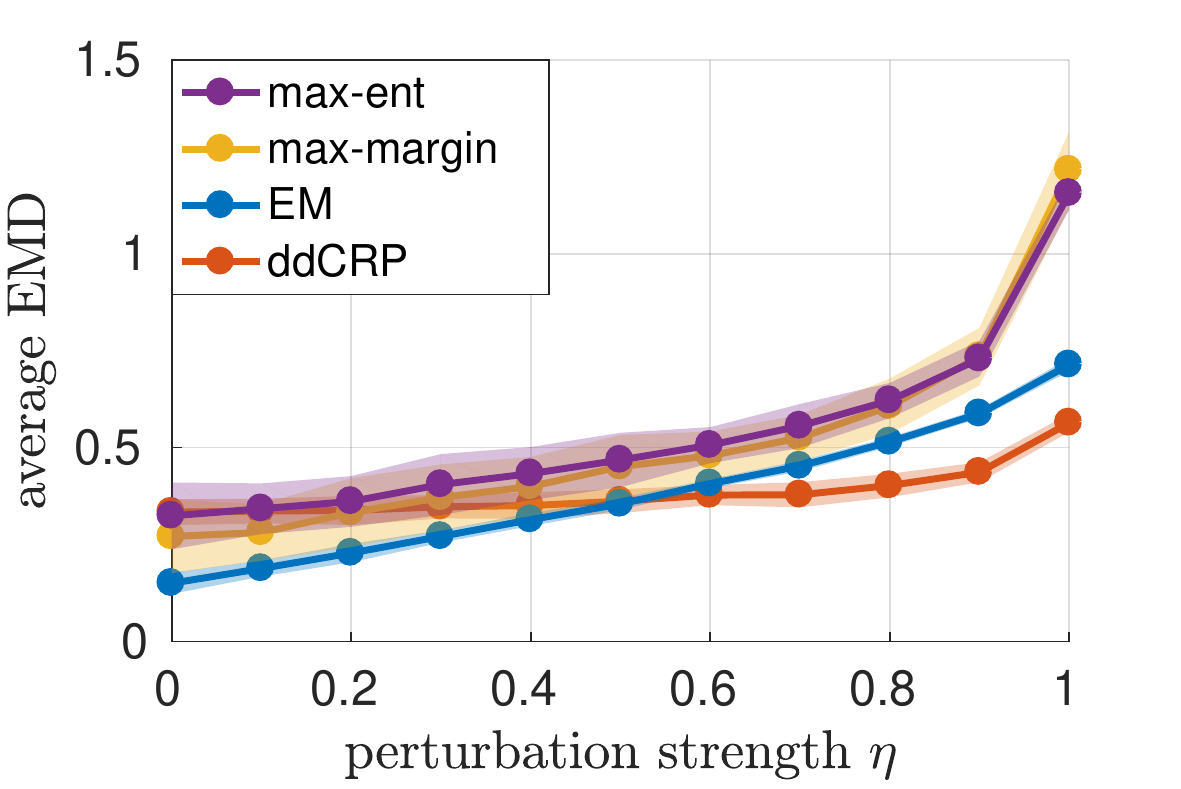}
		\caption{model robustness (MDP policy)}
		\label{subfig:modelingErrors}
	\end{subfigure}
	\caption{Average EMD values for the prediction task described in \cref{sec:discreteSetting}: (a) circular policy (b,c) MDP policy. Shown are the empirical mean values and standard deviations, estimated from 100 Monte Carlo runs. The EMD values are computed based on (a) the predicted action distributions and (b,c) the predicted next-state distributions. Note that the curves of max-ent (purple) and max-margin (yellow) in subfigure (a) lie on top of each other.}
\end{figure*}

In this section, we compare the prediction capabilities of our model to existing LfD frameworks, in particular: the maximum margin method in \cite{abbeel2004apprenticeship} (max-margin), the maximum entropy approach in \cite{ziebart2008maximum} (max-ent), and the expectation-maximization algorithm in \cite{sosic2016} (EM). For the comparison, we restrict ourselves to the ddCRP model 
which showed the best performance among all presented models. 

As a first experiment, we compare 
all methods on a 
finite version of the setting in \cref{sec:continuousExample}, which is obtained by discretizing the continuous state space into a regular grid $\mathcal{S}=\{(x,y)\in\mathbb{Z}^2:|x|,|y|\leq10\}$, resulting in a total of 441 states. The transition probabilities are chosen proportional to the normal densities in \cref{eq:transitionKernel} sampled at the grid points. Here, we used a noise level of $\sigma=1$ and a reduced number of eight actions. Probability mass ``lying outside'' the finite grid area is shifted to the closest border states of the grid. 

\Cref{subfig:circularGridWorld} delineates the average EMD over the number of trajectories (each of length $T=10$) provided for training. We observe that neither of the two intentional models (max-ent and max-margin) is able to capture the demonstrated expert behavior. This is due to the fact that the circular expert motion cannot be explained by a simple state-dependent reward structure but requires a more complex state-action reward model, which 
is not considered in the original formulations \cite{abbeel2004apprenticeship,ziebart2008maximum}. While the EM model is indeed able to capture the general trend of the data, the prediction is less accurate as compared to that of the ddCRP model, since it cannot reproduce the 
stochastic nature of the expert policy. In fact, 
this difference in performance will become even more pronounced 
for 
expert policies which distribute their probability mass on a larger subset of actions. Therefore, the ddCRP model 
outperforms all other models since 
the provided expert behavior violates their 
assumptions. 

To analyze how the ddCRP competes against the other models in their nominal situations, we further compare all algorithms on a standard grid world task where the expert behavior is obtained as the optimal response to a simple state-dependent reward function. 
Herein, each state on the grid is assigned a reward with a chance of 1\%, which is then drawn from a standard normal distribution. 
Worlds which contain no reward are discarded. The discount factor~0.9, which is used to compute the expert policy (see \cite{sutton1998reinforcement}), is provided as additional input for the intentional models. 
The results are shown in \Cref{subfig:MDPGridWorld}, which illustrates that the intention-based max-margin method outperforms all other methods for small amounts of training data. 
The sub-intentional methods (EM and ddCRP), on the other hand, yield better asymptotic estimates and smaller prediction variances. 
It should be pointed out that the three reference methods have a clear advantage over the ddCRP in this case 
because they assume a deterministic expert behavior \textit{a priori} and 
do not need to 
infer this piece of information from the data.
Despite 
this additional 
challenge, the ddCRP model yields a competitive performance. 

Finally, we compare all approaches in terms of their robustness against modeling errors. For this purpose, we repeat the previous experiment with a fixed number of 1000 trajectories but employ a different transition model for inference than used for data generation. More specifically, we utilize an overly fine-grained model consisting of 24 directions, assuming that the true action set is unknown, as suggested in \cref{sec:problemStatement}. Additionally, we perturb the assumed model by multiplying (and later renormalizing) each transition probability with a random number generated according to $f(u)=\tan(\frac{\pi}{4}(u+1))$, with $u\sim\textsc{Uniform}(-\eta,\eta)$ and perturbation strength $\eta\in[0,1]$. Due the resulting model mismatch, a comparison to the ground truth policy based on the predicted action distribution becomes meaningless. Instead, we compute the Euclidean EMDs between the true and the predicted next-state distributions, which we obtain by marginalizing the actions of the true/assumed transition model with respect to the true/learned policy. 
\Cref{subfig:modelingErrors} depicts the resulting prediction performance for different perturbation strengths~$\eta$. The results confirm that our approach is not only less sensitive to modeling errors as argued in \cref{sec:problemStatement}; also, the prediction variance is notably smaller than those of the intentional models.

\section{Conclusion}
\label{sec:conclusion}
In this work, we proposed a novel approach to 
LfD 
by jointly learning the latent control 
policy of an observed expert demonstrator together with 
a task-appropriate representation of the system state space.
%
%
With the 
described parametric and nonparametric models, 
we presented two formulations of the same problem that can be used either to learn a global system controller of specified complexity, or to 
infer the required model complexity 
from the observed 
expert behavior itself.
Simulation results for both 
countable and uncountable state spaces and a comparison to existing frameworks demonstrated the 
efficacy of our approach. Most notably, the results showed that our method is able to learn accurate predictive behavioral models in situations where intentional methods fail, \ie\ when the expert behavior cannot be explained as the 
result of a simple 
planning procedure. This makes our method applicable to a broader range of problems and suggests its use in a more general system identification context where we have no such prior knowledge about the expert behavior. 
Additionally, the 
task-adapted state representation learned through our framework can be used for further reasoning.

\bibliographystyle{IEEEtran}
\clearpage
\bibliography{bibliography.bib}
\vfill
\begin{IEEEbiography}[{\includegraphics[width=1in,height=1.25in,clip,keepaspectratio]{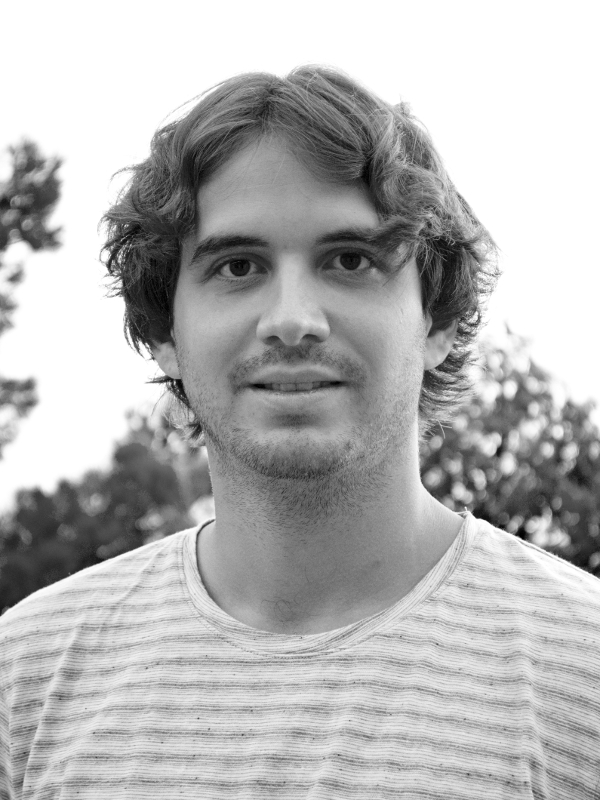}}]{Adrian \v{S}o\v{s}i\'{c}}
%
%
%
is a member of the Signal Processing Group and an associate member of 
the Bioinspired Communication Systems Lab 
at Technische Universit\"at Darmstadt. Currently, he is working towards his Ph.D.\ degree under the supervision of Prof.\ Abdelhak M.\ Zoubir and Prof.\ Heinz Koeppl. His research interests center around topics from machine learning and (inverse) reinforcement learning, with a focus on probabilistic inference, multi-agent systems, and Bayesian nonparametrics.

\end{IEEEbiography}
\vspace{-0.75cm}
\begin{IEEEbiography}[{\includegraphics[width=1in,height=1.25in,clip,keepaspectratio]{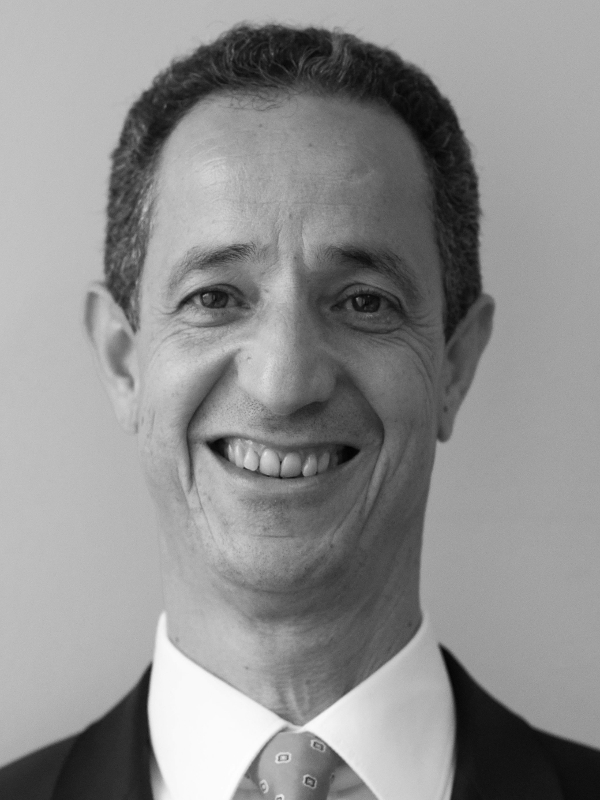}}]{Abdelhak M.\ Zoubir}
is professor at the Department of Electrical Engineering and Information Technology at Technische Universit\"at Darmstadt, Germany. His research interest lies in statistical methods for signal processing with emphasis on bootstrap techniques, robust detection and estimation, and array processing applied to telecommunications, radar, sonar, automotive monitoring and biomedicine.
\end{IEEEbiography}
\vspace{-0.75cm}
\begin{IEEEbiography}[{\includegraphics[width=1in,height=1.25in,clip,keepaspectratio]{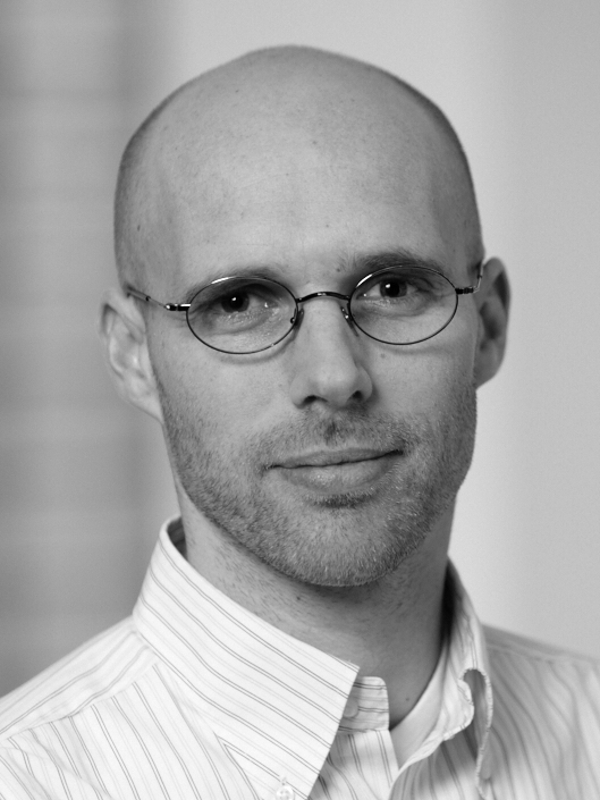}}]{Heinz Koeppl}
is professor at the Department of Electrical Engineering and Information Technology at Technische Universit\"at Darmstadt, Germany. His research interests include Bayesian inference methods for biomolecular data and methods for reconstructing large-scale biological or technological multi-agent systems from observational data.  

\end{IEEEbiography}


\section*{Supplementary material}

# A Bayesian Approach to Policy Recognition and State Representation Learning (Supplement)

Adrian Šošić, Abdelhak M. Zoubir and Heinz Koeppl

---

## A  Marginal Invariance & Policy Prediction in Large State Spaces

When we extended our reasoning to large state spaces in Section 2.3 using a reduced state space model (see Fig. S-1), we inevitably arrived at the following questions: By modeling the expert behavior only along observed trajectories, what does the resulting model imply for the remaining states of the state space? Can we still use it for predicting their local policies? The purpose of this section is to provide an in-depth discussion on the implications of this reduced modeling approach in the context of policy prediction.

When investigating the above-mentioned questions from a probabilistic perspective (i.e. by analyzing the induced joint distribution of our model), it turns out that they are strongly related to what is known as *marginal invariance* [1] (sometimes also referred to as *marginalization property* or simply *consistency* [2]). This property states that a model is consistent in the sense that it always provides the same marginal distributions for any subset of its variables, irrespective of the initial model size. In other words, a marginally invariant policy model yields the same answer for the given trajectory points, even if we include additional states into our reduced set $\widetilde{\mathcal{S}}$ for which we have not observed any demonstrations.

For our spatial models, that is, the Potts model and the ddCRP, it can be shown that this consistency property is indeed lacking (see [1] for a detailed discussion). This means that we cannot expect to get compatible results when conducting our reduced model inference on two data sets of different sizes. On the contrary, making predictions for new states would always require to rerun our Gibbs sampler on the augmented data set, including all additional states. This brings us to the following practical dilemma: imagine an on-line policy recognition scenario where we observe an expert controlling our system. After a certain period of time, we are asked to take over control, using the experience we have acquired during the observation period. Each control command, whether performed by the expert or by us, will trigger a new state transition, meaning that new data points arrive sequentially one after another. Consequently, it is impossible to decide in advance which states to include in

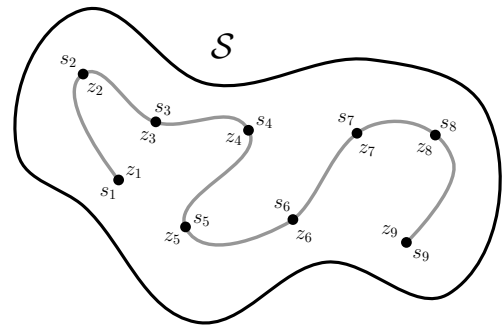

(a) reduced state space model

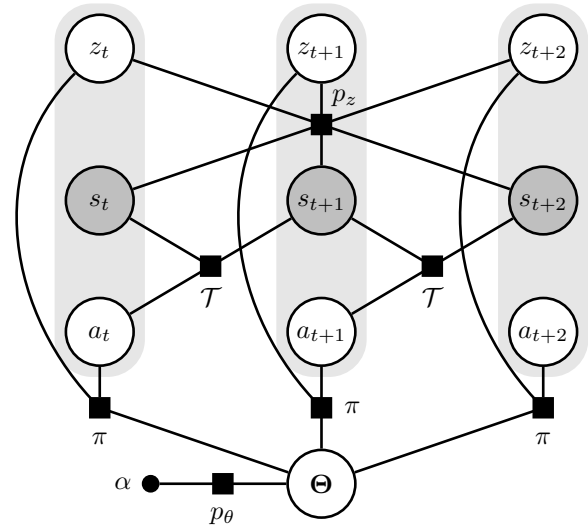

(b) corresponding factor graph

Fig. S-1: (a) Illustration of the reduced state space model, which operates on the space $\widetilde{\mathcal{S}} = \{s_1, s_2, \ldots, s_T\}$ of visited trajectory states. Note that the underlying decision-making process is assumed to be discrete in time; the continuous gray line shown in the figure is only to indicate the temporal ordering of the trajectory states. (b) Corresponding factor graph, highlighting the circular dependence between the variables. The factors are defined by the same building blocks that are used for the finite state space model. Observed variables are shaded in gray.

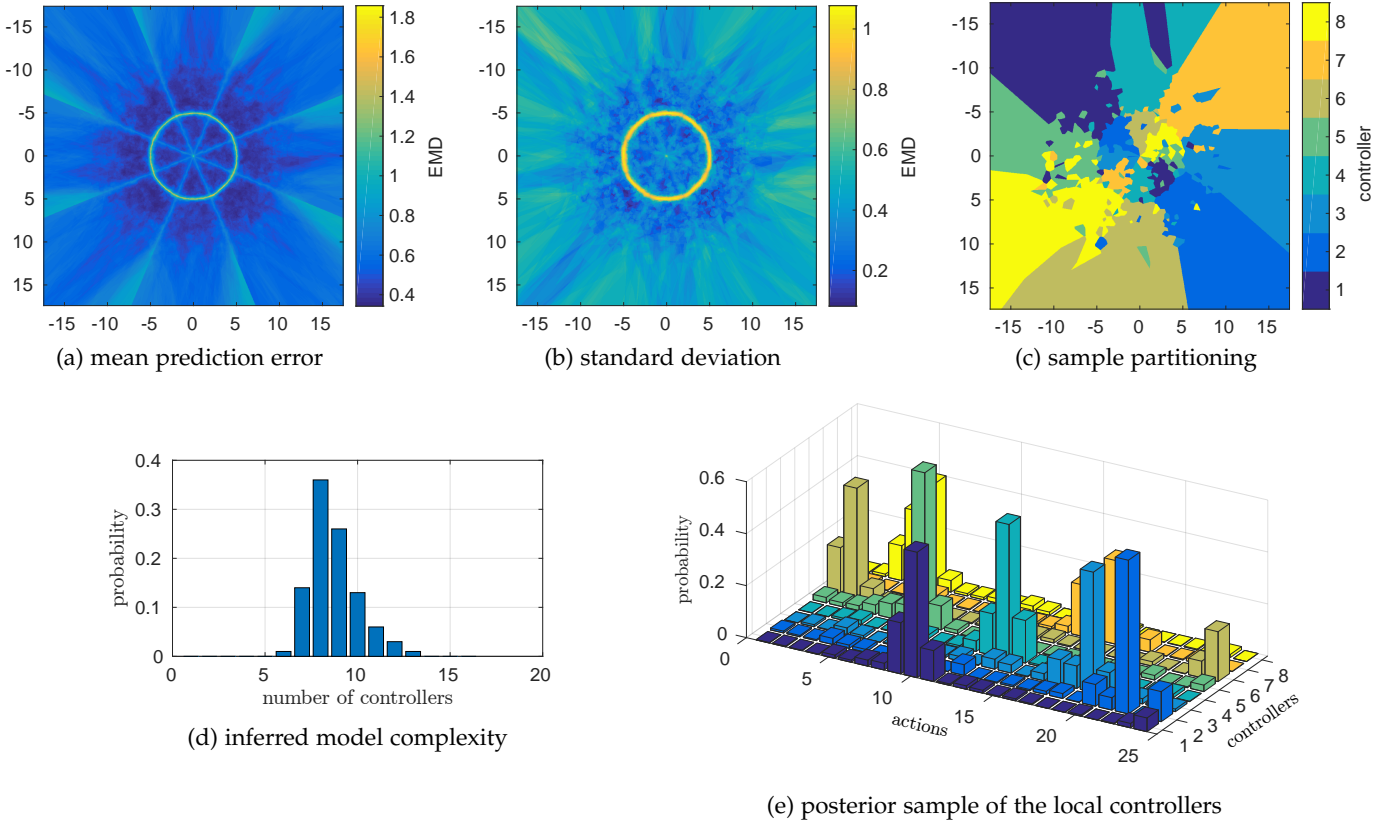

Fig. S-2: Simulation results for the ddCRP model on the continuous state space task described in Section 4.1. (a) Mean values of the spatial policy prediction error. (b) Standard deviations of the spatial policy prediction error. (c) Example partitioning of the state space, based on the local controllers depicted in sub-figure (e). The expert partitioning is shown in Fig. 4 of the main paper. (d) Posterior distribution of the number of local controllers. (e) Posterior sample of the local controllers found by the model. The results in (a,b,d) are based on 100 Monte Carlo runs while (c,e) are obtained from a single posterior sample. The figures in the top row were rendered using a spatial resolution of 2000x2000.

our reduced space $\widetilde{\mathcal{S}}$ and which not. A rigorous approach in the above-described sense would thus require to recalibrate the model after each state transition – a costly operation.

However, it is evident that the resulting data set is naturally divided into two disjoint parts, namely the expert demonstrations and the subsequent states reached during execution of the learned policy. Clearly, transitions occurring after the learning phase should by no means affect our belief about the expert policy and, hence, they should be completely discarded from the model. The easiest way to achieve this is, indeed, to "freeze" the model after the demonstration phase and to use the learned parameters to extrapolate the gathered policy information to surrounding states. This can be done, for instance, by retaining the structure of the involved spatial prior model to compute the resulting *maximum a posteriori* estimates for the extrapolated indicators of the new states, based on the inferred model parameters. In the case of the ddCRP, this coincides with the nearest-neighbor estimate (see Eq. (16)),

$$\hat{c}_{\text{new}} = \arg\max_{t \in \{1,\ldots,T\}} f(d_{t,\text{new}}) = \arg\min_{t \in \{1,\ldots,T\}} d_{t,\text{new}}. \quad \text{(S-1)}$$

Herein, $\hat{c}_{\text{new}}$ is the estimate for the indicator of the new state and $d_{t,\text{new}}$ denotes the distance of that state to the $t$th trajectory point.

Now, one could argue that the comfort of retaining a finite model structure for modeling inference problems on countably infinite or uncountable state spaces comes at the cost of not being able to provide a consistent posterior predictive distribution. However, the reduced state space approach allows us to incorporate the spatial information of the data in a fairly natural manner (i.e. in the form of pairwise distances), providing an easy way to model the expert behavior. Furthermore, our results demonstrate that the reduced model is able to capture the relevant spatial properties of a policy sufficiently accurate in order to make profound predictions about unseen states (see also subsequent section). Whether there exist alternative tractable models with similar properties remains to be seen.

## B  ADDITIONAL SIMULATION RESULTS

In this section, we provide additional simulation results for the ddCRP model on the continuous state space task described in Section 4.1.

Figure S-2a visualizes the spatial EMD prediction errors of the trained model in the form of a heat map, which compares the ground truth expert policy at non-trajectory points with the mean prediction provided by our model. The test points are placed on a regular grid of size 2000x2000 centered around the origin. The required indicator variables

at the interpolated states are computed according to Eq. (S-1). In line with our expectation, the prediction error reaches its maximum at the policy boundaries but is comparably small within each policy region, indicating a good model fit. Note that the "windmill shape" of the error can be explained as a result of the reduced state space approach in combination with the inherent asymmetry of the used data generation scheme: regions of the state space containing trajectory endings are locally underrepresented in the data set (see example trajectory in Fig. 4 in the paper); this increases the chance of assigning the end points of a trajectory to the cluster of the preceding region, resulting in a smearing of the previous cluster into the next region.

Also, we can observe that the variance of the error (Fig. S-2b) reaches its maximum at the transition regions and generally grows with the distance to the supporting trajectory data, reflecting the increasing prediction uncertainty at cluster boundaries and regions far from the expert demonstrations. Both figures were computed based on the learned policy representations of 100 Monte Carlo runs. Figure S-2c illustrates an example state partitioning of one such experiment, using the inferred local controllers depicted in Fig. S-2e. The result reveals that all expert motion patterns could be identified by our model. Note, however, that the two figures correspond to a single Gibbs sample of the process, which is *not* representative for the whole posterior distribution. Averaging over several experiments as done in Fig. S-2a and Fig. S-2b is not possible at the sample level due to the varying dimensionality of the corresponding policy representations (i.e. the number of learned controllers). Even taking averages over samples of equal dimensionality is not meaningful due to the multimodality of the posterior distribution, which arises from the inherent symmetry of the representation (i.e. interchanging two local controllers together with their corresponding indices yields the same model). Hence, averaging samples is possible only at the prediction level.

Finally, Fig. S-2d depicts the posterior distribution of the number of local controllers used by the model, which shows a pronounced peak at the true number used by the expert.

## C  COMPUTATIONAL COMPLEXITY

The overall computational cost of performing inference in our model depends largely on two factors: the complexity per Gibbs iteration and the mixing speed of the underlying Markov chain. Each Gibbs iteration consists of up to three stages: 1) sampling $T$ categorical action variables $\{a_t\}$ from the set $\{1, \ldots, |\mathcal{A}|\}$, where $T$ is the size of the demonstration set; 2) ddCRP model: sampling $N_S$ categorical state assignments $\{c_i\}$ from the set $\{1, \ldots, N_S\}$, where $N_S$ is the number of states (i.e. $|\mathcal{S}|$ or $|\widetilde{\mathcal{S}}|$); remaining models: sampling $N_S$ categorical partition assignments $\{z_i\}$ from the set $\{1, \ldots, K\}$, where $K$ is the number of local controllers; 3) for non-collapsed models: sampling $K$ Dirichlet-distributed control parameters $\{\boldsymbol{\theta}_k\}$ on the $(|\mathcal{A}| - 1)$-simplex.

Collapsing the control parameters generally improves the mixing speed of the chain (see Fig. 5 in the paper) but requires that action variables belonging to the same cluster be updated sequentially; hence, a non-collapsed strategy can be advantageous for larger data sets. Sampling the variables $\{a_t\}$, $\{\boldsymbol{\theta}_k\}$ and $\{z_i\}$ is computationally cheap because the involved action likelihoods $\{\mathcal{T}(s' \mid s, a)\}$ as well as the neighborhood structure $\mathcal{N}$ (Potts model) and the similarity values $\{f(d_{i,j})\}$ can be pre-computed. The most demanding operation is the update of $\{c_i\}$, which requires tracking the connected components of the underlying ddCRP graph. Using an appropriate graph representation, this can be done in polylogarithmic worst case time [3].


\end{document}